\documentclass[10pt,twocolumn,letterpaper]{article}

\usepackage{iccv}              
\usepackage{amssymb}
\usepackage{pifont}
\usepackage{multirow}
\usepackage{makecell}
\usepackage[accsupp]{axessibility}

\definecolor{iccvblue}{rgb}{0.21,0.49,0.74}
\usepackage[pagebackref,breaklinks,colorlinks,allcolors=iccvblue]{hyperref}

\def\paperID{10896} 
\def\confName{ICCV}
\def\confYear{2025}

\title{MotionLab: Unified Human Motion Generation and Editing via the Motion-Condition-Motion Paradigm}

\author{
    {Ziyan Guo{$^{1}$}} \quad Zeyu Hu{$^{2}$} \quad De Wen Soh{$^{1}$} \quad Na Zhao$^{1}$\footnotemark[2] \\
	\small
	$^{1}$\, Singapore University of Technology and Design, Singapore ~~ $^{2}$\ LIGHTSPEED, Singapore \\ 
	\small
	{\tt\small ziyan\_guo@mymail.sutd.edu.sg, \{dewen\_soh, na\_zhao\}@sutd.edu.sg, hzy9724@gmail.com}
	}

\begin{document}
\twocolumn[{%
\maketitle
\begin{center}
  \centering
  \captionsetup{type=figure}
  \vspace{-8mm}
  \includegraphics[width=\linewidth,page=1]{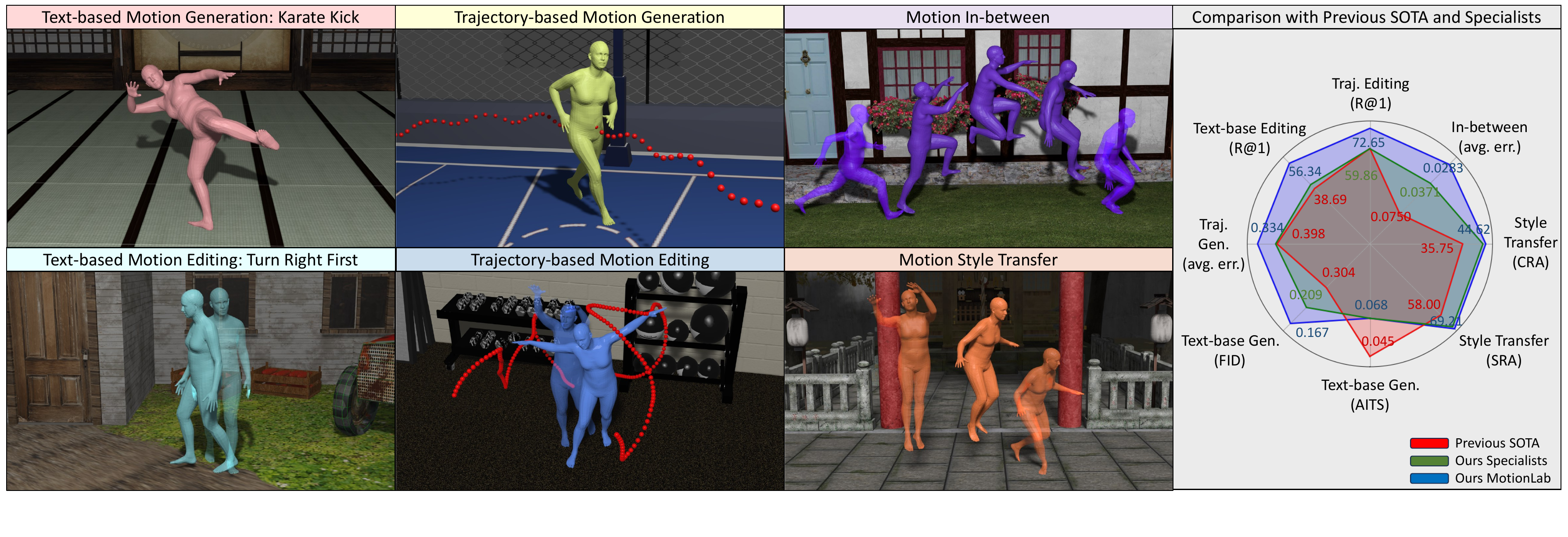}
  \vspace{-10mm}
  \caption{Demonstration of our MotionLab's versatility, performance and efficiency. Ours specialists refer to the proposed framework tailored for specified tasks. Previous SOTA refer to multiple models, including MotionLCM \cite{dai2025motionlcm}, OmniControl \cite{xie2023omnicontrol}, MotionFix \cite{athanasiou2024motionfix}, CondMDI \cite{cohan2024flexible} and MCM-LDM \cite{Song_2024_CVPR}. All motions are represented using SMPL \cite{loper2023smpl}, where transparent motion indicates the source motion or condition, and the other represents the target motion. \textbf{More qualitative results are available in the website and appendix.}}
  \label{fig:demo}
\end{center} 
}]
\renewcommand{\thefootnote}{\fnsymbol{footnote}}
\footnotetext[2]{Corresponding author: Na Zhao.}

\begin{abstract}
Human motion generation and editing are key components of computer vision. However, current approaches in this field tend to offer isolated solutions tailored to specific tasks, which can be inefficient and impractical for real-world applications. While some efforts have aimed to unify motion-related tasks, these methods simply use different modalities as conditions to guide motion generation. Consequently, they lack editing capabilities, fine-grained control, and fail to facilitate knowledge sharing across tasks. 
To address these limitations and provide a versatile, unified framework capable of handling both human motion generation and editing, we introduce a novel paradigm: \textbf{Motion-Condition-Motion}, which enables the unified formulation of diverse tasks with three concepts: source motion, condition, and target motion.
Based on this paradigm, we propose a unified framework, \textbf{MotionLab}, which incorporates rectified flows to learn the mapping from source motion to target motion, guided by the specified conditions.
In MotionLab, we introduce the 1) MotionFlow Transformer to enhance conditional generation and editing without task-specific modules; 2) Aligned Rotational Position Encoding to guarantee the time synchronization between source motion and target motion; 3) Task Specified Instruction Modulation; and 4) Motion Curriculum Learning for effective multi-task learning and knowledge sharing across tasks.
Notably, our MotionLab demonstrates promising generalization capabilities and inference efficiency across multiple benchmarks for human motion.
Our code and additional video results are available at: \href{https://diouo.github.io/motionlab.github.io/}{\textcolor{blue}{https://diouo.github.io/motionlab.github.io/}}.
\end{abstract}

\vspace{-5mm}
\section{Introduction}
\label{sec:intro}

Human motion is a crucial component of computer vision, with applications spanning game development, film production, and virtual reality \cite{Guo_2022_CVPR,tevet2023human}. 
With the advancements of generative diffusion models \cite{ho2020denoising,song2020denoising,dhariwal2021diffusion}, human motion generation has garnered considerable attention, aiming at generating human motion aligned with the input conditions, such as text \cite{tevet2023human,zhang2022motiondiffuse,tevet2022motionclip} and trajectory (\ie, joints' coordinates) \cite{xie2023omnicontrol,fujiwara2025chronologically,sun2024lgtm,dai2025motionlcm,athanasiou2023sinc,athanasiou2022teach,zhang2023finemogen,song2023loss,sampieri2024length}. Concurrently, to maximize the utility of motion assets within industry settings, significant efforts have been dedicated to motion editing tasks, including motion style transfer \cite{zhong2025smoodi,Song_2024_CVPR,aberman2020unpaired, guo2024generative, jang2022motion}.

\begin{table*}[!t]
\resizebox{\linewidth}{!}{
\begin{tabular}{ccccccc}
\hline
Method & text-based generation & text-based editing & trajectory-based generation & trajectory-based editing & in-between & style transfer\\ 
\hline
MDM \cite{tevet2023human} & $\checkmark$ &$\times$ &$\times$ &$\times$ & $-$ &$\times$ \\  
MLD \cite{chen2023executing} &$\checkmark$ &$\times$ &$\times$ &$\times$ &$\times$ &$\times$ \\ 
OmniControl \cite{xie2023omnicontrol} &$\checkmark$ &$\times$ &$\checkmark$ &$\times$ & $-$ &$\times$  \\  
MotionFix \cite{athanasiou2024motionfix} & $-$ &$\checkmark$ &$\times$ &$\times$  & $-$ &$\times$ \\ 
CondMDI \cite{cohan2024flexible} &$\checkmark$ &$\times$ &$\checkmark$ &$\times$  &$\checkmark$ &$\times$ \\
MCM-LDM \cite{Song_2024_CVPR} &$\times$ &$\times$ &$\times$ &$\times$  & $-$ &$\checkmark$ \\ 
MotionGPT \cite{jiang2023motiongpt} &$\checkmark$ & $-$ &$\times$  &$\times$ &$\checkmark$ &$\times$ \\ 
MotionCLR \cite{chen2024motionclr} & $\checkmark$ &$-$ &$\times$ &$\times$ & $-$ &$-$ \\
\hline
Ours &$\checkmark$ &$\checkmark$ &$\checkmark$ &$\checkmark$ &$\checkmark$ &$\checkmark$ \\ \hline  
\end{tabular}
}
\caption{Summary of methods focusing on motion generation and editing. $\checkmark$ indicates that the method has been trained for the task, $\times$ indicates that the method fails to implement, and $-$ indicates that the method has not been trained but can implement in a zero-shot manner.}
\label{tab:comparison}
\vspace{-3mm}
\end{table*}

As summarized in Table~\ref{tab:comparison}, current research in this domain mainly develops task-specific solutions, forcing practitioners to train multiple models for human motion generation and editing, which is inefficient and impractical. Although several studies \cite{shrestha2025generating,zhou2023ude,zhou2023unified,zhang2025large,yang2024unimumo,luo2024m,fan2024everything2motion} have attempted to unify motion-related tasks, they merely consider different modalities as generation conditions, leading to limited editing capabilities and insufficient fine-grained trajectory control. Moreover, these approaches overlook the intrinsic links between motion generation and editing, thereby hindering potential knowledge sharing. In contrast, a well-designed unified framework can exploit the large volumes of multi-task data to surpass specialist models through effective cross-task representation learning. Motivated by this prospect and inspired by the success of large language models in unifying NLP tasks \cite{achiam2023gpt,dubey2024llama}, we pose the following question: \textit{\textbf{Can human motion generation and editing be effectively unified within a single framework?}}

In response to this question, it is essential to design an elegant and scalable paradigm.
Hence, we propose a novel paradigm: \textbf{Motion-Condition-Motion}. This paradigm is built upon three concepts -- \textit{source motion}, \textit{condition}, and \textit{target motion}. 
Concretely, the target motion is predicted by the source motion and specified conditions in this Motion-Condition-Motion paradigm.
For any human motion generation task, the source motion can be treated as none, and the target motion must align with the provided conditions. For any human motion editing task, the target motion is derived from the source motion based on the conditions. By unifying these tasks within this elegant and scalable paradigm, this framework can be seamlessly extended to various human motion tasks and scaled across diverse datasets. 
Given that human motions are inherently tied to their semantics, trajectories, and styles in practical applications, we aim to unify several key tasks under this framework. These tasks include \textit{text-based motion generation and editing} \cite{tevet2023human,zhang2022motiondiffuse,tevet2022motionclip, athanasiou2024motionfix, goel2024iterative}, \textit{trajectory-based motion generation and editing} \cite{xie2023omnicontrol,dai2025motionlcm, guo2025tstmotion}, \textit{motion in-between} \cite{cohan2024flexible,harvey2020robust} and \textit{motion style transfer} \cite{Song_2024_CVPR,zhong2025smoodi}, as illustrated in Figure~\ref{fig:demo}.

Despite the proposed paradigm, several significant challenges remain in balancing versatility, performance, and efficiency: 
1) Unifying various tasks inevitably introduces additional modalities, while each modality may involve multiple tasks. A naive solution, like adopting multiple cross-attention mechanisms for each task in generation-unified frameworks \cite{fan2024everything2motion,zhang2025large}, is suboptimal. 
2) More sampling time is required for certain tasks (\eg, trajectory-based motion generation and motion in-between \cite{xie2023omnicontrol,zhong2025smoodi}), as existing methods in these areas involve task-specific posterior guidance \cite{chung2022diffusion} during inference to improve conditional guidance.
3) Time asynchrony between the source motion and target motion may arise due to the limited scale of the paired editing dataset and the use of implicit positional encoding \cite{chen2023executing, xie2023omnicontrol,athanasiou2024motionfix,cohan2024flexible}.
4) Most importantly, naively integrating various motion generation and editing tasks into a single framework could lead to task conflicts and catastrophic forgetting, impairing the framework's overall performance.

To address these challenges, we propose a novel generative framework, termed \textbf{MotionLab}, built upon rectified flows \cite{liu2022flow,lipman2022flow} and MM-DiT \cite{esser2024scaling}, as illustrated in Figure~\ref{fig:illustration}. Rectified flows are particularly well-suited for MotionLab since they are designed to implement the optimal transport between source and target distributions, naturally aligning with the Motion-Condition-Motion paradigm. Furthermore, human motion must adhere to skeletal kinematics. Consequently, the distribution of valid target motions is highly restricted, and individual target motions may correspond to multiple conditions (\textit{i.e.}, many-to-one), which suggests that a shared transport map can be efficiently transferred across tasks.
In contrast to MM-DiT, our proposed Motion Flow Transformer (MFT) encompasses more modalities, including source motion, target motion, text, trajectory, and style. Within the MFT, each modality is assigned a dedicated path, and comprehensive interaction among modalities is facilitated via joint attention. This architecture enables MFT to advance conditional generation and editing capabilities without necessitating task-specific modules or posterior guidance for particular tasks.
To ensure temporal synchronization between source and target motions, we incorporate an Aligned Rotational Position Encoding into MFT, which explicitly aligns tokens at corresponding frames between the source and target sequences. Moreover, to enable adaptation of a single modality to various tasks, we introduce Task Instruction Modulation, which flexibly embeds various tasks into the MFT. To seamlessly integrate diverse tasks, we propose a curriculum-inspired training strategy, termed Motion Curriculum Learning, based on an easy-to-hard training principle. 

In this paper, motion generation and editing are decomposed into combinations of modalities through the Motion-Condition-Motion paradigm. These modalities are subsequently represented via each modality's paths within the MFT, learning inter-modal interactions through the joint attention, while adapting individual modalities to different tasks through Task Instruction Modulation. By implementing a curriculum learning from single to multiple, from simple (\eg, source motion and trajectory) to complex (\eg, text and style) modalities, the spatial knowledge inherent in 3D representations can be effectively transferred to the latter modalities since the modalities of the former can represent the latter.
Through these designs, we validate MotionLab on multiple benchmarks, demonstrating superior versatility, performance, and efficiency compared to baselines across various human motion generation and editing tasks.

\section{Related Work}
\textbf{Motion Generation and Editing.} Motion generation can be classified based on input conditions. Among these, text-based motion generation is one of the most compelling areas \cite{zhang2022motiondiffuse,tevet2022motionclip,tevet2023human,Guo_2022_CVPR,chen2023executing,guo2024momask,guo2022tm2t,wang2024motiongpt,jiang2023motiongpt,kim2023flame,lin2023motion,lu2023humantomato,plappert2016kit,zhang2023generating,guo2020action2motion,petrovich2021action}, as it trains models to comprehend the semantics of text and generate corresponding pose sequences. To address the fine-grained requirements of practical applications, trajectory-based motion generation has been proposed \cite{karunratanakul2023guided,shafir2023human,xie2023omnicontrol,dai2025motionlcm,zhang2023finemogen}, where specific motion properties, such as joints reaching designated positions at specified times, are defined. Additionally, motion in-between \cite{tevet2023human,jiang2023motiongpt,cohan2024flexible,qin2022motion,pinyoanuntapong2024mmm} focuses on generating complete motion sequences given key poses at keyframes. To enable in-place editing of human motion \cite{goel2024iterative,athanasiou2024motionfix}, MotionFix \cite{athanasiou2024motionfix} introduces text-based motion editing using paired source and target motions. We extend this approach to trajectory-based motion editing by substituting text with joint trajectories. Meanwhile, style plays a crucial role in human motion, leading to motion style-transfer \cite{jang2022motion,aberman2020unpaired,zhong2025smoodi,Song_2024_CVPR}. However, the aforementioned methods concentrate solely on specific tasks, rendering them impractical for real-world applications. Moreover, they overlook the intrinsic connections across different human motion tasks and fail to facilitate knowledge sharing among these tasks. In contrast, our unified framework enhances performance on data-scarce editing tasks through multi-task learning.

\vspace{0.02in}
\noindent\textbf{Unified Frameworks for Human Motion.} There are also some efforts in existing methods that try to unify tasks related to human motion. One line of work \cite{jiang2023motiongpt,jiang2025motionchain,zhou2024avatargpt,li2024unimotion,wang2024motiongpt,ling2024motionllama,wu2024motionllm,luo2024m,athanasiou2024motionfix} focuses on motion understanding, such as motion captioning or describing human motion in images and videos. Yet, these approaches often rely on GPT-like structures, which require a large amount of training resources and GPU memory. In addition, they fail to provide fine-grained control (\eg, trajectory-based generation and editing) over motion, which is crucial in practical applications. Another line of effort \cite{shrestha2025generating,zhou2023ude,zhou2023unified,zhang2025large,yang2024unimumo,luo2024m,fan2024everything2motion,alexanderson2023listen} highlights generating motion based on more modalities, such as music and speech. However, these approaches only integrate more modalities into one model and cannot flexibly edit motion, which can cause them to suffer from multi-task learning and limit their scope of use. The closest to our work are FLAME \cite{kim2023flame} and MotionCLR \cite{chen2024motionclr}.
However, FLAME does not support style transfer and precise text-based editing like ``move faster'', and MotionCLR does not support trajectory-based generation and editing, requiring cumbersome manual adjustments to the attention.
\section{Preliminary: Rectified Flows}

\label{sec:preliminaries}
Flow-based generative methods \cite{liu2022flow,lipman2022flow,esser2024scaling,ma2024sit,fei2024flux,polyak2024moviegencastmedia} have recently received significant attention due to their generalizability and efficiency compared to diffusion models. Specifically,
these methods directly regress the transport vector field between the source distribution $p_1$ and target distribution $p_0$ with the straightest possible trajectories and sample by the corresponding ordinary differential equation (ODE) \cite{wang2024frieren}. Among these methods, rectified flows \cite{liu2022flow,lipman2022flow} aim to learn a trajectory from source data $x_0$ to target data $x_1$, which can be formulated as $x_t = \varphi(x_0,x_1,t)$, and the velocity field $v_t$ of the trajectory $x_t$ can be defined by:
\begin{align}
& v_t = \frac{dx_t}{dt} = \frac{\partial\varphi_t(x_0,x_1,t)}{\partial t}, t\in[0,1]
\end{align}
Once we have learned this velocity field $v_t$, we can get $x_0$ from any $x_1$ by numerically integrating:
\begin{align}
& x_{t-\frac{1}{N}} = x_t - \frac{1}{N}v_\theta(t,x_t)
\end{align}
where $N$ is the discretization number of the interval [0,1]. Hence, rectified flows $v_\theta$ are trained to predict $v_t$ by given $x_t$ and $t$, and the training objective can be represented as:
\begin{align}
&\mathcal L_{RF}(\theta) = \int^1_0 \mathbb{E}_{(x_0,x_1)\sim(p_0,p_1)}[||v_\theta(t, x_t)-v_t||^2_2]dt
\end{align}

\section{Motion-Condition-Motion} \label{sec:mcm}
To unify the tasks of human motion generation and editing, we propose the paradigm of Motion-Condition-Motion. As shown in Table~\ref{tab:paradigm}, all these tasks are unified by three concepts: \textit{source motion}, \textit{condition}, and \textit{target motion}. 

\noindent\textbf{Motion Generation}. For the motion generation tasks, including \textit{text}/\textit{trajectory-based} generation and \textit{motion in-between}, the source motion can be treated as none, with the target motion aligning to the corresponding conditions. For instance, in \textit{text-based generation}, the generated motion should align with the semantics of the provided text, such as ``karate kick'' illustrated in Figure~\ref{fig:demo}. \textit{Masked reconstruction}, as a specific motion generation task, requires the target motion to align with the masked source motion in the specified frames without relying on additional conditions.
Notably, the \textit{unconditional generation} (given zero frames) and \textit{reconstruction} (given all frames) are special cases of masked reconstruction, thus these three tasks can share the same task instruction as described in Section~\ref{sec:tsim}.

\noindent\textbf{Motion Editing}. For motion editing, the source motion must be provided, and the target motion is derived from the source motion based on the specified conditions. In the case of \textit{text-based motion editing}, the generated motion should originate from the source motion, with modifications applied only to the specified parts as dictated by the provided text, such as ``use the opposite leg''. For \textit{trajectory-based editing}, the source motion should be aligned with the given joints' coordinates, ensuring that the specified joints in the source motion are accurately moved to the designated positions within the specified frames. In \textit{motion style transfer}, the generated motion should adopt the style of the style motion while preserving the semantics of the source motion.

\noindent\textbf{Remarks.} In particular, \textit{trajectory-based motion generation} and \textit{motion in-between} are highly similar, as they both aim to ensure that specific joints reach designated positions at specific times. 
Their primary difference is that the former is sparse in space (\ie, joints) but dense in time, whereas the latter is dense in space (\ie, joints) but sparse in time. To efficiently share the parameters and learned representations between the two tasks, we unify their conditions into a single condition. Meanwhile, \textit{masked reconstruction} is also similar to these two tasks. However, while these two tasks only include the coordinates of joints, the source motion also encompasses the velocity and angular velocity of joints. Therefore, they represent different modalities, and masked reconstruction constitutes a distinct task.

\section{MotionLab}
\label{sec:method}

\begin{figure*}[!t]
  \centering
  \includegraphics[width=\linewidth,page=1]{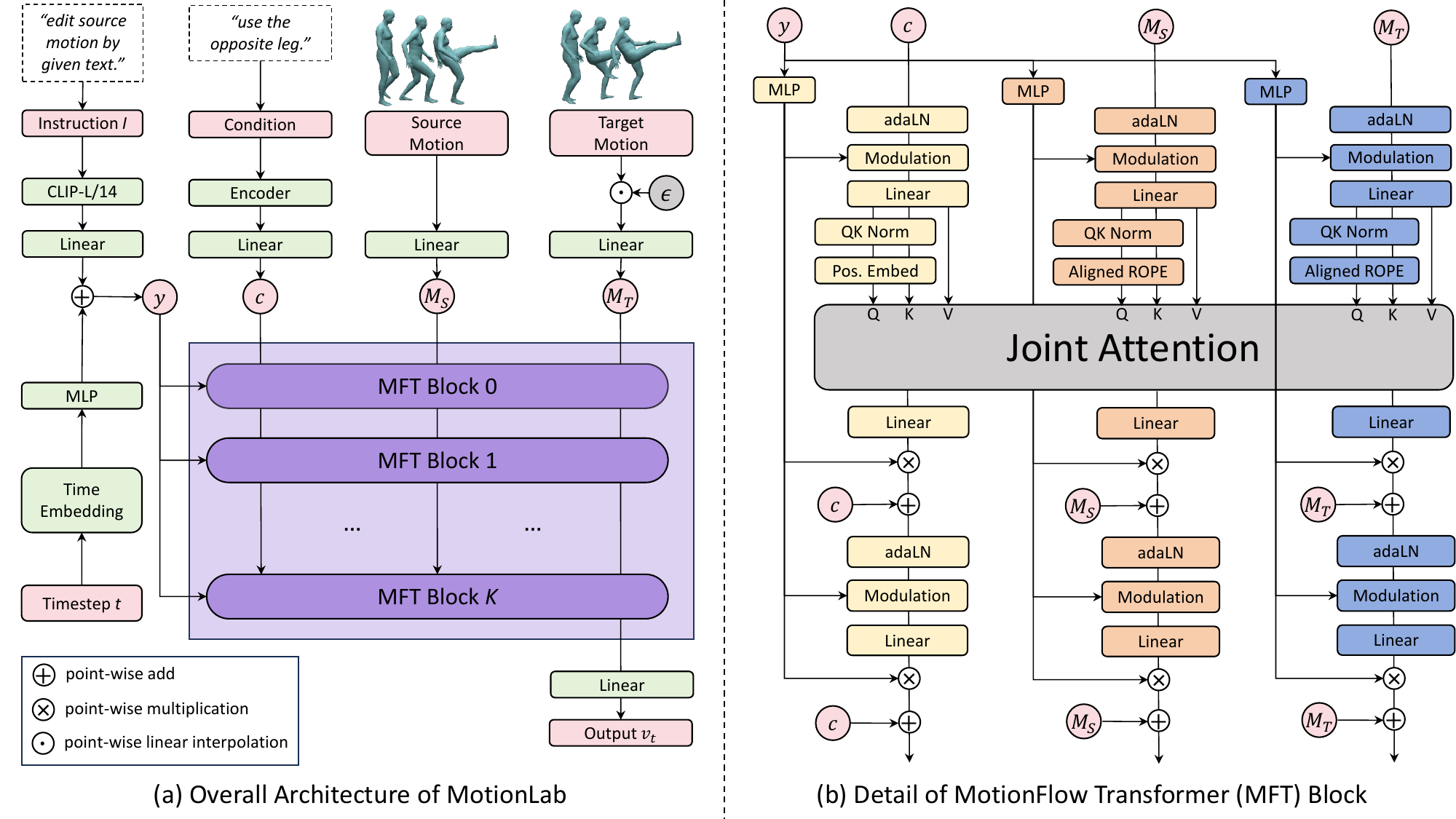}
  \vspace{-5mm}
  \caption{Illustration of our MotionLab and the detail of its MotionFlow Transformer (MFT).}
  \label{fig:illustration}
\end{figure*} 

\begin{table}[!t]
\resizebox{\linewidth}{!}{
\begin{tabular}{cccc}
\hline
Task & Source Motion & Condition & Target Motion\\ \hline
unconditional generation  & $\emptyset$ &$\emptyset$ & $\checkmark$ \\
masked reconstruction & masked source motion & $\emptyset$ & source motion \\
reconstruction & complete source motion & $\emptyset$ & source motion \\
\hline
text-based generation & $\emptyset$ &text & $\checkmark$ \\
trajectory-based generation & $\emptyset$ &text/joints' coordinates & $\checkmark$ \\
motion in-between &  $\emptyset$ &text/poses in keyframes & $\checkmark$ \\ \hline
text-based editing & $\checkmark$ &text & $\checkmark$ \\
trajectory-based editing & $\checkmark$ &text/joints' coordinates & $\checkmark$ \\
style transfer  & $\checkmark$ &style motion & $\checkmark$ \\ 
\hline
\end{tabular}
}
\caption{Structuring human motion tasks within our Motion-Condition-Motion paradigm.}
\vspace{-5mm}
\label{tab:paradigm}
\end{table}

Based on our proposed Motion-Condition-Motion paradigm, we introduce a unified framework named \textbf{MotionLab}, as illustrated in Figure~\ref{fig:illustration}(a). The core of MotionLab is the \textbf{MotionFlow Transformer (MFT)} (Sec.~\ref{sec:mft}), inspired by MM-DiT \cite{esser2024scaling}, which leverages rectified flow to map source motion $M_S \in \mathbb{R}^{N_{S}\times D}$ to target motion $M_T \in \mathbb{R}^{N_{T}\times D}$ based on the corresponding condition $C$ for each task.

To enable task differentiation, we propose \textbf{Task Instruction Modulation} (Sec.~\ref{sec:tsim}), where a task-specific instruction $I\in\mathbb{R}^{1\times768}$ extracted from the CLIP \cite{radford2021learning} is also input into MFT alongside $M_S$, $M_T$, and $C$. At each timestep $t$, MFT is trained to predict velocity field $v_t$, which is derived via linear interpolation between target motion $M_T$ and Gaussian noise $\epsilon \in \mathbb{R}^{N_{T}\times D}$. 

For effective multi-task training, we adopt \textbf{Motion Curriculum Learning} (Sec.~\ref{sec:mcl}) which organizes tasks hierarchically to facilitate learning. Once trained, MotionLab can map $M_S$ to $M_T$ based on the specified $C$, by predicting $v_t$ in descending order of timestep $t$ as described in Sec.~\ref{sec:preliminaries}.

\vspace{-1mm}
\subsection{MotionFlow Transformer} \label{sec:mft}
As shown in the Figure~\ref{fig:illustration} (b), MotionFlow Transformer contains three key components: \textit{Joint Attention} to interact tokens from different modalities; \textit{Modality Path} for distinguishing tokens from different modalities and extracting their representations, and \textit{Aligned ROPE} for position encoding of modalities with time information.

\noindent\textbf{Joint Attention.} 
We first adopt
the joint attention mechanism \cite{esser2024scaling}, through which tokens from different modalities can interplay with each other. Specifically, all these tokens will be projected to the query, key, and value representations, and then will be concatenated into a sequence of orderly tokens. Subsequently, these orderly tokens are applied by the attention operation, whose output is again split into corresponding tokens of different modalities.

\noindent\textbf{Modality Path.} While the joint attention can interact with tokens from different modalities, there is still a need to differentiate between different tokens. In addition to the QKV projection and FeedForward Network (FFN) in the attention mechanism, 
as used in MM-DiT, our MFT incorporates the adaptive Layer Normalization (adaLN) and a modulation mechanism \cite{peebles2023scalable} for each modality, 
enhancing conditional generation and editing capabilities.

\noindent\textbf{Aligned Rotational Position Encoding.} 
Considering that the use of absolute position encoding in existing methods \cite{chen2023executing} can weaken the temporal alignment between source motion and target motion due to the limited scale of paired datasets, we adopt a relative position encoding method, ROtational Position Encoding (ROPE) \cite{su2024roformer}. ROPE explicitly embeds the relative distances between tokens, preserving temporal relationships more effectively. Instead of naively applying a 3-dimensional ROPE to distinguish source motion, target motion, and conditions with time information (\eg, trajectory), we propose Aligned ROPE, which encodes these components with appropriate temporal information using a 1-dimensional ROPE. This design avoids the confusion caused by 3-dimensional ROPE, where distances between tokens within a modality can interfere with cross-modality distances, ensuring better temporal alignment.

\subsection{Task Instruction Modulation} \label{sec:tsim}
MM-DiT implements a 
modulation mechanism that enhances text-to-image generation through the incorporation of textual embeddings (\eg, ``a photo of a dog") as modulation signals. However, within our unified framework, various tasks necessitate the integration of multiple modalities, and critically, identical modalities may require distinct representational forms across different tasks. This complexity renders approaches such as learned task tokens (\eg, [TASK]) or one-hot encoding vectors inadequate for managing arbitrary numbers and combinations of modalities.

Recognizing the inherent flexibility of natural language, we leverage textual representations acquired by foundation models (\eg, CLIP) to effectively differentiate identical modalities across disparate tasks. For instance, we utilize the textual embedding of ``\textit{edit source motion by given style}" to facilitate the adaptation of source motion to style transfer. This approach, while conceptually straightforward, provides remarkable effectiveness in enhancing system flexibility and scalability, thereby enabling seamless extension to diverse tasks involving multiple modalities.

\subsection{Motion Curriculum Learning} \label{sec:mcl}
To achieve effective multi-task learning and facilitate knowledge sharing between tasks, we propose an easy-to-hard hierarchical training strategy inspired by curriculum learning \cite{bengio2009curriculum}. Specifically, new tasks are sequentially introduced into the training based on their difficulty, guided by the following assumptions: 1) The fewer modalities a task involves, the simpler the task; 2) Editing tasks are easier than generating tasks, as only the conditional difference between source motion and target motion needs to be learned; 3) The more specific the conditional information (\eg, source motion) provided, the simpler the task becomes. The importance of these three criteria decreases in order. Guided by the easy-to-hard training principle, the training process in MotionLab is divided into two stages: \textit{self-supervised pre-training} and \textit{supervised fine-tuning}.

\noindent\textbf{Pre-training.} Intuitively, the reconstruction of masked source motion is the easiest task. Hence, we first train the model based on the masked source motion, independent of the conditions. This approach allows the model to learn prior motion representations independent of conditions, thereby generalizing to different tasks. Following MoMask \cite{guo2024momask}, we randomly mask from zero frames to all frames. This flexible strategy provides tasks of varying difficulty levels, avoiding overfitting on simple tasks (all frames) and mode collapse on difficult tasks (zero frames). Furthermore, this strategy seamlessly performs source motion reconstruction (\ie, all frames) and unconditional training (\ie, zero frames), which is crucial for Classifier-Free Guidance (CFG) \cite{ho2022classifier}. 
Unlike MoMask, which masks all joints in a single frame simultaneously due to its discrete tokens, we extend masked pre-training to randomly mask joint trajectories to enhance the understanding of in-between and trajectory-based tasks.
Specifically, we pre-train MotionLab using these three tasks (\ie, masked source motion reconstruction, trajectory-based generation without text, and in-between without text) for 1,000 epochs.

\noindent\textbf{Fine-tuning.} 
In the supervised fine-tuning stage, we train MotionLab on tasks in an easy-to-hard sequence. Specifically, a new task is introduced into training every 200 epochs in the following order: \ding{192} text-based generation, \ding{193} style-based generation (an auxiliary task for training the modality path of the style, not our primary goal), \ding{194} trajectory-based editing (without text), \ding{195} text-based editing, \ding{196} style transfer,  \ding{197} motion in-between and trajectory-based generation, \ding{198} trajectory-based editing. 
This progressive learning strategy ensures effective adaptation and knowledge sharing across tasks.
Particularly, \ding{192} and \ding{193} are the simplest tasks because they only include one modality, whereas others include at least two modalities. Among tasks involving two modalities, \ding{194}, \ding{195}, and \ding{196} take priority over \ding{197} since they are editing tasks. Additionally, as text is less specific than trajectory but more specific than style,
the order is \ding{194}, \ding{195}, and \ding{196}.

To mitigate catastrophic forgetting, previous tasks are trained with new tasks, based on the probability derived from the FID of the last evaluation. However, the FID scales for different tasks vary due to their differing difficulty levels. Consequently, we use the percentage change compared to the previous evaluation as the probability, which encourages the model to re-learn forgotten tasks or tasks that it has not yet fully mastered. To support classifier-free guidance, we also train the model to unconditionally generate and reconstruct the complete source motion. Empirically, in this stage, a 5\% probability is allocated for unconditional generation, 5\% for reconstructing the complete source motion, 45\% for previous tasks, and 45\% for the new task.


In summary, this training strategy has many advantages: 1) it enables our framework to adapt to various tasks; 2) it seamlessly supports CFG during inference; 3) it allows flexible management of the training process to avoid retraining due to errors. Meanwhile, this training strategy, from single modality to multiple modalities, can be considered as first learning the representation of each modality separately and then learning the representation of the interaction between multiple modalities, which can be distinguished by the Task Instruction Modulation. 
Furthermore, by prioritizing the introduction of spatial conditions (\ie, source motion and trajectory), this strategy can share the model's understanding between them and abstract conditions (\ie, text and style), as the latter conditions can be represented by the former.

\begin{table}[]
\resizebox{\linewidth}{!}{
\begin{tabular}{ccccccc}
\hline
Method &  FID$\downarrow$ & R@3$\uparrow$  & Diversity$\rightarrow$ & \makecell{MM \\Dist$\downarrow$} & MModality$\uparrow$ & AITS$\downarrow$ \\ \hline
GT & 0.002 & 0.797 & 9.503 & 2.974 & 2.799 & - \\ \hline
T2M \cite{Guo_2022_CVPR} & 1.087 & 0.736 & 9.188 & 3.340 & 2.090 & \textbf{0.040} \\
MDM \cite{tevet2023human} & 0.544 & 0.611 & \textbf{9.559} & 5.566 & \underline{2.799} & 26.04 \\
MotionDiffuse \cite{zhang2022motiondiffuse} & 1.954 & 0.739 & 11.10  & 2.958 & 0.730 & 15.51 \\
MLD \cite{chen2023executing} & 0.473 & 0.772 & 9.724  & 3.196 & 2.413 & 0.236 \\
T2M-GPT\cite{zhang2023generating} & \textbf{0.116} & 0.775 & 9.761 & 3.118 & 1.856 & 1.124 \\ 
MotionGPT \cite{jiang2023motiongpt} & 0.232 & 0.778 & \underline{9.528}  & 3.096 & 2.008 & 1.240 \\
CondMDI \cite{cohan2024flexible} & 0.254 & 0.6450 & 9.749 & - & - & 57.25 \\
MotionLCM \cite{dai2025motionlcm} & 0.304 & 0.698 & 9.607 & 3.012 & 2.259 & \underline{0.045} \\
MotionCLR \cite{chen2024motionclr} & 0.269 & \textbf{0.831} & 9.607 & \textbf{2.806} & 1.985 & 0.830 \\
\hline
\textbf{Ours} & \underline{0.167} & \underline{0.810} & 9.593 & \underline{2.830} & \textbf{2.912} & 0.068 \\
\hline
\end{tabular}
}
\caption{Evaluation of \textit{text-based motion generation} on HumanML3D \cite{Guo_2022_CVPR} dataset. The models in bold are the optimal models, and the models in underline are the sub-optimal models. }
\label{tab:text}
\end{table}

\begin{table}[]
\resizebox{\linewidth}{!}{
\begin{tabular}{cccccccc}
\hline
Method & Joints & FID$\downarrow$ & R@3$\uparrow$  & Diversity$\rightarrow$ & \makecell{Foot skate\\ratio$\downarrow$} & \makecell{Average\\Error$\downarrow$} & AITS$\downarrow$ \\ \hline
GT & - & 0.002 & 0.797 & 9.503 & 0.000 & - & - \\ \hline
GMD \cite{karunratanakul2023guided} & pelvis & 0.576 & 0.665 & 9.206 & 0.101 & \underline{0.1439} & 137.0 \\
PriorMDM \cite{shafir2023human} & pelvis & 0.475 & 0.583 & 9.156 & - & 0.4417 & 19.83 \\
OmniControl \cite{xie2023omnicontrol} & pelvis & \underline{0.212} & 0.678 & 9.773 & \underline{0.057} & 0.3226 & 39.78 \\
MotionLCM \cite{dai2025motionlcm} & pelvis & 0.531 & \textbf{0.752} & \underline{9.253} & - & 0.1897 & \textbf{0.035} \\
\textbf{Ours} & pelvis & \textbf{0.095} & \underline{0.740} & \textbf{9.502} & \textbf{0.007} & \textbf{0.0286} & \underline{0.133} \\ \hline
OmniControl  \cite{xie2023omnicontrol} & all & 0.310 & 0.693 & 9.502 & 0.061 & 0.0404 & 76.71 \\
\textbf{Ours} & all & 0.126 & 0.765  & 9.554 & 0.002 & 0.0334 & 0.134 \\
\hline
\end{tabular}
}
\caption{Evaluation of \textit{trajectory-based motion generation} on HumanML3D \cite{Guo_2022_CVPR} dataset.}
\label{tab:trajectory}
\end{table}

\begin{table}[]
\resizebox{\linewidth}{!}{
\begin{tabular}{ccccccccc}
\hline
\multirow{2}{*}{Method} & \multirow{2}{*}{Condition} & \multicolumn{4}{c}{generated-to-target retrieval} & \multirow{2}{*}{\makecell{Average\\Error$\downarrow$} } &  \multirow{2}{*}{AITS $\downarrow$}  \\ \
& & R@1$\uparrow$ & R@2$\uparrow$ & R@3$\uparrow$ & AvgR $\downarrow$ & &   \\ \hline
GT  & - & 100.0 & 100.0 & 100.0 & 1.00 & - & - \\ \hline                     
TMED$^*$ \cite{athanasiou2024motionfix} & text  & 38.69 & 50.61 & 62.23 & 4.15 & - & 26.57 \\ 
\textbf{Ours} & text  & 56.34 & 70.40 & 77.24 & 3.54 & - & 0.16 \\ \hline

TMED$^*$ \cite{athanasiou2024motionfix} & trajectory  & 60.01 & 73.33 & 82.69 & 2.67 & 0.129 & 30.56 \\ 
\textbf{Ours} & trajectory  & 72.65 & 82.71 & 87.89 & 2.20 & 0.027 & 0.19 \\ \hline
\end{tabular}
}
\caption{Evaluation of \textit{text-based and trajectory-based motion editing} on MotionFix \cite{athanasiou2024motionfix} dataset. TMED$^*$ mean that we re-implement the models since the original models are in the skeleton of SMPL format, while ours is in HumanML3D format.}
\label{tab:edit}
\end{table}

\begin{figure*}[!h]
  \centering
  \includegraphics[width=\linewidth]{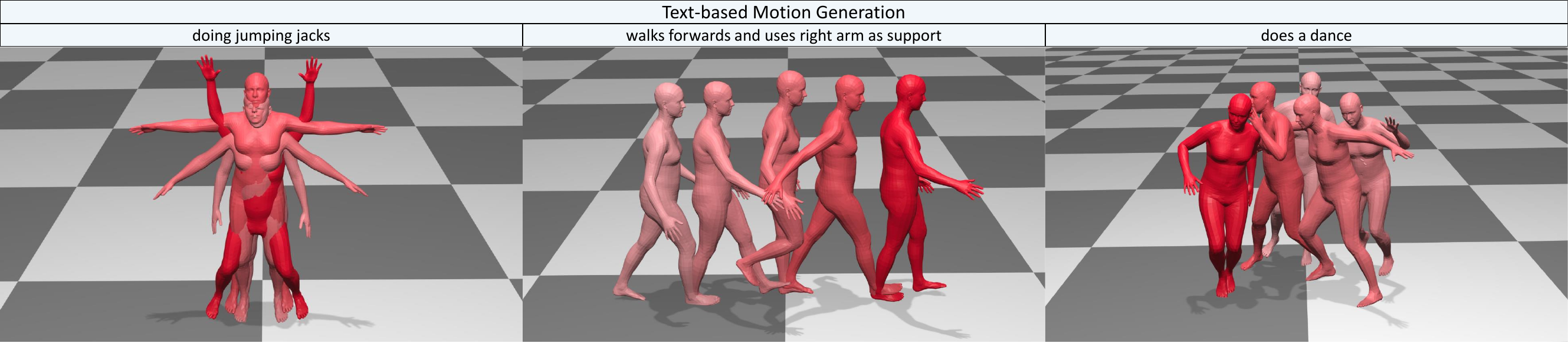}
  \vspace{-6mm}
  \caption{Qualitative results on the text-based motion generation. For clarity, as time progresses, motions transit from light to dark colors.}
  \label{fig:quantitive_text}
  \vspace{-2mm}
\end{figure*} 

\begin{figure*}[!h]
  \centering
  \includegraphics[width=\linewidth]{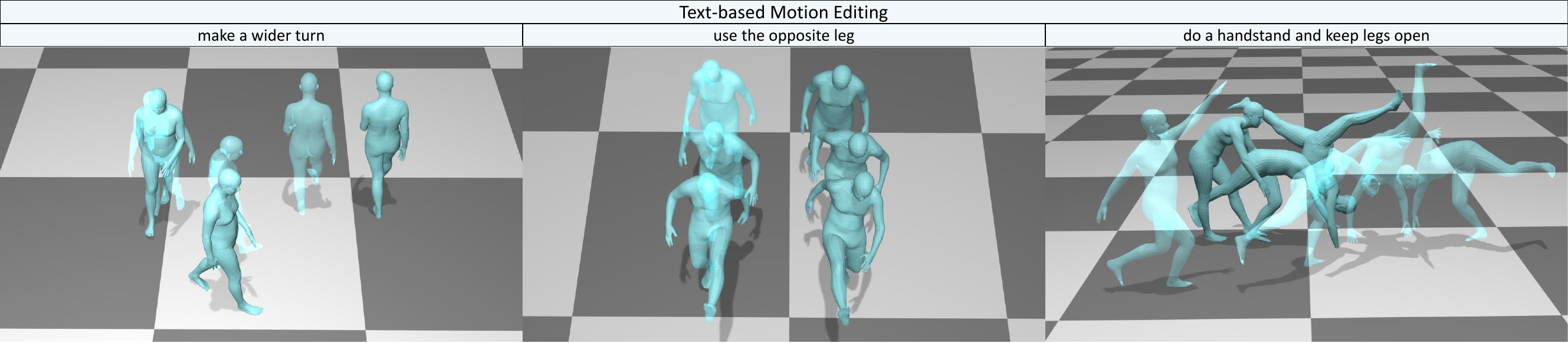}
    \vspace{-6mm}
  \caption{Qualitative results on text-based motion editing. The transparent motion is source motion, and the other is the generated motion.}
  \label{fig:quantitive_edit}
    \vspace{-2mm}
\end{figure*} 

\begin{figure*}[!h]
  \centering
  \includegraphics[width=\linewidth]{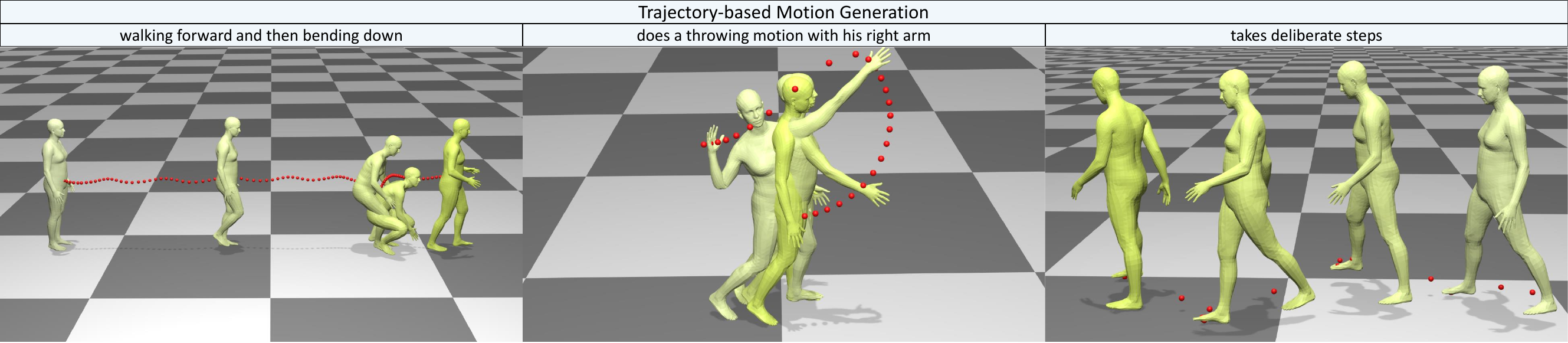}
    \vspace{-6mm}
  \caption{Qualitative results on the trajectory-based motion generation. The red balls are the trajectory of the pelvis, right hand and foot.}
  \label{fig:quantitive_trajectory}
    \vspace{-4mm}
\end{figure*} 

\section{Experiments}
\label{sec:experiments}

\textbf{Datasets.} 
To evaluate the text-based motion generation, the trajectory-based motion generation, motion in-between, and motion style transfer, we leverage the HumanML3D \cite{Guo_2022_CVPR} dataset, which comprises 14,646 motions and 44,970 motion annotations. To evaluate the text-based and trajectory-based motion editing, we utilize MotionFix \cite{athanasiou2024motionfix} dataset, which is the first dataset for text-based human motion editing, including 6,730 motion pairs. 

\vspace{0.05in}
\noindent\textbf{Evaluation Metrics.} 
We evaluate our framework using the following metrics: 1) To evaluate \textit{text-based motion generation}, following the \cite{chen2023executing}, we adopt the FID to evaluate the distribution gap between the generated and original motions; Diversity to calculate the corresponding variance between motions; R-precision (R@K) to measure the proximity of the generated motion to the text or motion; Foot skating ratio to evaluate the physical plausibility of motion; Multi-modal Distance (MM Dist) calculates the distance between motions and texts. We also introduce Average Inference Time per Sample (AITS) measured in seconds to evaluate the inference efficiency; 2) To evaluate \textit{trajectory-based motion generation} and \textit{motion in-between}, following \cite{xie2023omnicontrol}, we adopt the Average Error to measures the mean distance between the generated motion locations and the keyframe locations; 3) To evaluate \textit{text-based and trajectory-based motion editing}, following \cite{athanasiou2024motionfix}, we adopt the AvgR to measure the success rate of retrieval from edited motion to target motion; 4) To evaluate \textit{motion style transfer}, following \cite{Song_2024_CVPR}, we adopt the Style Recognition Accuracy (SRA) and Content Recognition Accuracy (CRA) to measure the stylistic and content accuracy of the generated motion; Trajectory Similarity Index (TSI) to evaluate the trajectory preservation from source motion.

\begin{table*}[!t]
\resizebox{\linewidth}{!}{
\begin{tabular}{cccccccc}
\hline
Method & text gen. (FID) & traj. gen. (avg. err.) & text edit (R@1) & traj. edit (R@1) & in-between (avg. err.) & style transfer (CRA) & style transfer (SRA) \\ \hline
w/o rectified flows & 0.301 & 0.0359 & 54.38 & 69.21 & 0.0289 & 42.20 & 63.96 \\
w/o MotionFlow Transformer & 0.483 & 0.0447 & 51.26 & 65.34 & 0.0349 & 35.36 & 53.83 \\
w/o Aligned ROPE & 0.253 & 0.0886 & 45.39 & 61.99 & 0.0756 & 42.23 & 56.59 \\
w/o task instruction modulation & 0.223 & 0.0401 & 55.96 & 70.01 & 0.0288 & 40.55 & 63.91 \\
w/o motion curriculum learning & 1.956 & 0.1983 & 28.56 & 36.61 & 0.1682 & 29.51 & 34.23 \\
\hline
Ours specialist models & \underline{0.209} & \underline{0.0398} & \underline{41.44} & \underline{59.86} & \underline{0.0371} & \underline{43.53} & \underline{67.55} \\
\textbf{Ours} & \textbf{0.167} & \textbf{0.0334} & \textbf{56.34} & \textbf{72.65} & \textbf{0.0283} & \textbf{44.62} & \textbf{69.21} \\
\hline
\end{tabular}
}
\vspace{-.1in}
\caption{
\small{\textbf{Ablation studies of key components of MotionLab on each task}. Refer to the text for the detailed configuration of each variant.}
}
\label{tab:ablation}
\end{table*}

\noindent\textbf{Implementation Details.} 
In order to fairly compare our model with other models, motions from all datasets have been retargeted into one skeleton following HumanML3D format with 20 fps, where the number of joints $J$ is 22, and the dimension of motion feature $D$ is 263. The learning rate is set to be 1$\times$10$^{-4}$. The timesteps are set to 1,000 for training and 50 for inference. Our models are trained by four RTX 4090D with each batch of 64 for 4 days. To ensure a fair comparison, the AITS of all models are recalculated using one RTX 4090D.

\subsection{Quantitative Results}

\textbf{Overall Performance.} 
As shown in Table~\ref{tab:text} to Table~\ref{tab:edit}, MotionLab demonstrates promising performance across \textit{all benchmarks}\footnote[1]{Due to space limitations, we include the quantitative results on motion in-between and motion style transfer in the supplementary material.}, underscoring the effectiveness of our framework's design. Notably, as MotionLab is a unified framework without task-specific designs, it must balance versatility, performance, and efficiency. 

Specifically, as shown in Table~\ref{tab:text} and Figure~\ref{fig:timestep}, MotionLab achieves superior performance (lowest FID, which is the key metric for generation tasks) with relatively fast inference time (third-lowest AITS). For trajectory-based tasks (Table~\ref{tab:trajectory}) and the motion in-between task, MotionLab achieves lower average error. We believe these improvements stem from the effectiveness of masked pre-training and Aligned ROPE, which ensures spatial and temporal synchronization between the trajectory and target motion. 

\subsection{Qualitative Results}

As shown in the Figure~\ref{fig:quantitive_text} from Figure~\ref{fig:quantitive_trajectory}, our framework presents its powerful capabilities to generate motion aligned with the conditions and edit source motion based on the condition, demonstrating its versatility and performance. For more visualization results, please kindly refer to the supplementary and project website.

\subsection{Ablation Studies}
We perform several ablation experiments\footnote[1]{Additional ablation studies are available in the supp. material.} on our framework to validate the designs in MotionLab and report the results in Table~\ref{tab:ablation}: the $1^{st}$ variant replaces rectified flows with diffusion models; the $2^{nd}$ variant uses a regular transformer (\ie, without modulation mechanism and adopting cross-attention) instead of MFT. The $3^{rd}$ variant uses the implicit 1D-learnable encoding instead of Aligned ROPE; The $4^{th}$ variant does not adopt the Task Instruction Modulation; the $5^{th}$ variant directly learns all tasks based on their FID compared to the last evaluation. 
Additionally, we 
use the same model to train specialist models for each task, denoted as `our specialist models' in Table~\ref{tab:ablation}.

As can be seen from the results, the removal of motion curriculum learning markedly diminishes model performance across all tasks, underscoring its pivotal role in facilitating knowledge transfer between tasks. Meanwhile, our unified framework outperforms our specialist models in all tasks, potentially due to the knowledge sharing of motion curriculum learning. These phenomena can also be attributed to the strategy's capacity to enable the model to integrate its comprehension of spatial conditions (\eg, source motion, trajectory, and intermediate states) with abstract conditions (\eg, text and style), given that the latter can be partially represented by the former. 
Furthermore, as shown in Table~\ref{tab:ablation},
 Aligned ROPE is essential for space-related tasks, significantly reducing the average error. It effectively aligns source motion and target motion temporally, contributing to high R-precision in editing tasks.

\begin{figure}[!t]
  \centering
  \includegraphics[width=0.98\linewidth]{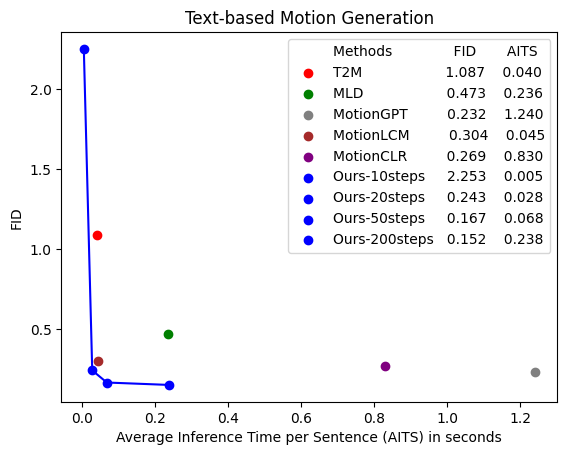}
  \vspace{-.05in}
  \caption{\small{
  \textbf{Impact of timesteps during inference on MotionLab}. The closer to the lower left corner, the more powerful the model.}}
   \vspace{-.1in}
  \label{fig:timestep}
\end{figure} 


Additionally, we evaluate the impact of timesteps during inference on our MotionLab and compare its performance with baseline methods in terms of generation quality and inference time for the text-based motion generation task. As shown in Figure~\ref{fig:timestep}, our framework strikes an optimal balance between generation quality and efficiency.

\section{Conclusion}
Building on our proposed Motion-Condition-Motion paradigm, we have developed the MotionLab framework to unify human motion generation and editing. We have introduced the MotionFlow Transformer to leverage rectified flows to learn the mapping from source motion to target motion based on specified conditions. Additionally, we have incorporated Aligned Rotational Position Encoding to ensure synchronization between source motion and target motion, Task Instruction Modulation, and Motion Curriculum Learning for effective multi-task learning. Our proposed MotionLab framework demonstrates superior versatility, performance, and efficiency compared to existing state-of-the-art methods and our specialist models.

\vspace{0.05in}
\noindent\textbf{Acknowledgments.} 
This research is supported by the Ministry of Education, Singapore, under its MOE Academic Research Fund Tier 1 -- SMU-SUTD Internal Research Grant (SMU-SUTD 2023\_02\_09).
\newpage
{\small
    \bibliographystyle{ieeenat_fullname}
    \bibliography{main}

\begin{thebibliography}{74}
\providecommand{\natexlab}[1]{#1}
\providecommand{\url}[1]{\texttt{#1}}
\expandafter\ifx\csname urlstyle\endcsname\relax
  \providecommand{\doi}[1]{doi: #1}\else
  \providecommand{\doi}{doi: \begingroup \urlstyle{rm}\Url}\fi

\bibitem[Aberman et~al.(2020)Aberman, Weng, Lischinski, Cohen-Or, and Chen]{aberman2020unpaired}
Kfir Aberman, Yijia Weng, Dani Lischinski, Daniel Cohen-Or, and Baoquan Chen.
\newblock Unpaired motion style transfer from video to animation.
\newblock \emph{ACM Transactions on Graphics (TOG)}, 39\penalty0 (4):\penalty0 64--1, 2020.

\bibitem[Achiam et~al.(2023)Achiam, Adler, Agarwal, Ahmad, Akkaya, Aleman, Almeida, Altenschmidt, Altman, Anadkat, et~al.]{achiam2023gpt}
Josh Achiam, Steven Adler, Sandhini Agarwal, Lama Ahmad, Ilge Akkaya, Florencia~Leoni Aleman, Diogo Almeida, Janko Altenschmidt, Sam Altman, Shyamal Anadkat, et~al.
\newblock Gpt-4 technical report.
\newblock \emph{arXiv preprint arXiv:2303.08774}, 2023.

\bibitem[Alexanderson et~al.(2023)Alexanderson, Nagy, Beskow, and Henter]{alexanderson2023listen}
Simon Alexanderson, Rajmund Nagy, Jonas Beskow, and Gustav~Eje Henter.
\newblock Listen, denoise, action! audio-driven motion synthesis with diffusion models.
\newblock \emph{ACM Transactions on Graphics (TOG)}, 42\penalty0 (4):\penalty0 1--20, 2023.

\bibitem[Athanasiou et~al.(2022)Athanasiou, Petrovich, Black, and Varol]{athanasiou2022teach}
Nikos Athanasiou, Mathis Petrovich, Michael~J Black, and G{\"u}l Varol.
\newblock Teach: Temporal action composition for 3d humans.
\newblock In \emph{2022 International Conference on 3D Vision (3DV)}, pages 414--423. IEEE, 2022.

\bibitem[Athanasiou et~al.(2023)Athanasiou, Petrovich, Black, and Varol]{athanasiou2023sinc}
Nikos Athanasiou, Mathis Petrovich, Michael~J Black, and G{\"u}l Varol.
\newblock Sinc: Spatial composition of 3d human motions for simultaneous action generation.
\newblock In \emph{Proceedings of the IEEE/CVF International Conference on Computer Vision}, pages 9984--9995, 2023.

\bibitem[Athanasiou et~al.(2024)Athanasiou, Cseke, Diomataris, Black, and Varol]{athanasiou2024motionfix}
Nikos Athanasiou, Alp{\'a}r Cseke, Markos Diomataris, Michael~J Black, and G{\"u}l Varol.
\newblock Motionfix: Text-driven 3d human motion editing.
\newblock In \emph{SIGGRAPH Asia 2024 Conference Papers}, pages 1--11, 2024.

\bibitem[Bengio et~al.(2009)Bengio, Louradour, Collobert, and Weston]{bengio2009curriculum}
Yoshua Bengio, J{\'e}r{\^o}me Louradour, Ronan Collobert, and Jason Weston.
\newblock Curriculum learning.
\newblock In \emph{Proceedings of the 26th annual international conference on machine learning}, pages 41--48, 2009.

\bibitem[Chen et~al.(2024)Chen, Dai, Ju, Lu, and Zhang]{chen2024motionclr}
Ling-Hao Chen, Wenxun Dai, Xuan Ju, Shunlin Lu, and Lei Zhang.
\newblock Motionclr: Motion generation and training-free editing via understanding attention mechanisms.
\newblock \emph{arXiv preprint arXiv:2410.18977}, 2024.

\bibitem[Chen et~al.(2023)Chen, Jiang, Liu, Huang, Fu, Chen, and Yu]{chen2023executing}
Xin Chen, Biao Jiang, Wen Liu, Zilong Huang, Bin Fu, Tao Chen, and Gang Yu.
\newblock Executing your commands via motion diffusion in latent space.
\newblock In \emph{Proceedings of the IEEE/CVF Conference on Computer Vision and Pattern Recognition}, pages 18000--18010, 2023.

\bibitem[Chung et~al.(2022)Chung, Kim, Mccann, Klasky, and Ye]{chung2022diffusion}
Hyungjin Chung, Jeongsol Kim, Michael~T Mccann, Marc~L Klasky, and Jong~Chul Ye.
\newblock Diffusion posterior sampling for general noisy inverse problems.
\newblock \emph{arXiv preprint arXiv:2209.14687}, 2022.

\bibitem[Cohan et~al.(2024)Cohan, Tevet, Reda, Peng, and van~de Panne]{cohan2024flexible}
Setareh Cohan, Guy Tevet, Daniele Reda, Xue~Bin Peng, and Michiel van~de Panne.
\newblock Flexible motion in-betweening with diffusion models.
\newblock In \emph{ACM SIGGRAPH 2024 Conference Papers}, pages 1--9, 2024.

\bibitem[Dai et~al.(2025)Dai, Chen, Wang, Liu, Dai, and Tang]{dai2025motionlcm}
Wenxun Dai, Ling-Hao Chen, Jingbo Wang, Jinpeng Liu, Bo Dai, and Yansong Tang.
\newblock Motionlcm: Real-time controllable motion generation via latent consistency model.
\newblock In \emph{European Conference on Computer Vision}, pages 390--408. Springer, 2025.

\bibitem[Dhariwal and Nichol(2021)]{dhariwal2021diffusion}
Prafulla Dhariwal and Alexander Nichol.
\newblock Diffusion models beat gans on image synthesis.
\newblock \emph{Advances in neural information processing systems}, 34:\penalty0 8780--8794, 2021.

\bibitem[Dubey et~al.(2024)Dubey, Jauhri, Pandey, Kadian, Al-Dahle, Letman, Mathur, Schelten, Yang, Fan, et~al.]{dubey2024llama}
Abhimanyu Dubey, Abhinav Jauhri, Abhinav Pandey, Abhishek Kadian, Ahmad Al-Dahle, Aiesha Letman, Akhil Mathur, Alan Schelten, Amy Yang, Angela Fan, et~al.
\newblock The llama 3 herd of models.
\newblock \emph{arXiv preprint arXiv:2407.21783}, 2024.

\bibitem[Esser et~al.(2024)Esser, Kulal, Blattmann, Entezari, M{\"u}ller, Saini, Levi, Lorenz, Sauer, Boesel, et~al.]{esser2024scaling}
Patrick Esser, Sumith Kulal, Andreas Blattmann, Rahim Entezari, Jonas M{\"u}ller, Harry Saini, Yam Levi, Dominik Lorenz, Axel Sauer, Frederic Boesel, et~al.
\newblock Scaling rectified flow transformers for high-resolution image synthesis.
\newblock In \emph{Forty-first International Conference on Machine Learning}, 2024.

\bibitem[Fan et~al.(2024)Fan, Ji, Xu, Shen, and Chen]{fan2024everything2motion}
Zhaoxin Fan, Longbin Ji, Pengxin Xu, Fan Shen, and Kai Chen.
\newblock Everything2motion: Synchronizing diverse inputs via a unified framework for human motion synthesis.
\newblock In \emph{Proceedings of the AAAI Conference on Artificial Intelligence}, pages 1688--1697, 2024.

\bibitem[Fei et~al.(2024)Fei, Fan, Yu, and Huang]{fei2024flux}
Zhengcong Fei, Mingyuan Fan, Changqian Yu, and Junshi Huang.
\newblock Flux that plays music.
\newblock \emph{arXiv preprint arXiv:2409.00587}, 2024.

\bibitem[Fujiwara et~al.(2025)Fujiwara, Tanaka, and Yu]{fujiwara2025chronologically}
Kent Fujiwara, Mikihiro Tanaka, and Qing Yu.
\newblock Chronologically accurate retrieval for temporal grounding of motion-language models.
\newblock In \emph{European Conference on Computer Vision}, pages 323--339. Springer, 2025.

\bibitem[Goel et~al.(2024)Goel, Wang, Liu, and Fatahalian]{goel2024iterative}
Purvi Goel, Kuan-Chieh Wang, C~Karen Liu, and Kayvon Fatahalian.
\newblock Iterative motion editing with natural language.
\newblock In \emph{ACM SIGGRAPH 2024 Conference Papers}, pages 1--9, 2024.

\bibitem[Guo et~al.(2020)Guo, Zuo, Wang, Zou, Sun, Deng, Gong, and Cheng]{guo2020action2motion}
Chuan Guo, Xinxin Zuo, Sen Wang, Shihao Zou, Qingyao Sun, Annan Deng, Minglun Gong, and Li Cheng.
\newblock Action2motion: Conditioned generation of 3d human motions.
\newblock In \emph{Proceedings of the 28th ACM International Conference on Multimedia}, pages 2021--2029, 2020.

\bibitem[Guo et~al.(2022{\natexlab{a}})Guo, Zou, Zuo, Wang, Ji, Li, and Cheng]{Guo_2022_CVPR}
Chuan Guo, Shihao Zou, Xinxin Zuo, Sen Wang, Wei Ji, Xingyu Li, and Li Cheng.
\newblock Generating diverse and natural 3d human motions from text.
\newblock In \emph{Proceedings of the IEEE/CVF Conference on Computer Vision and Pattern Recognition (CVPR)}, pages 5152--5161, 2022{\natexlab{a}}.

\bibitem[Guo et~al.(2022{\natexlab{b}})Guo, Zuo, Wang, and Cheng]{guo2022tm2t}
Chuan Guo, Xinxin Zuo, Sen Wang, and Li Cheng.
\newblock Tm2t: Stochastic and tokenized modeling for the reciprocal generation of 3d human motions and texts.
\newblock In \emph{European Conference on Computer Vision}, pages 580--597. Springer, 2022{\natexlab{b}}.

\bibitem[Guo et~al.(2024{\natexlab{a}})Guo, Mu, Javed, Wang, and Cheng]{guo2024momask}
Chuan Guo, Yuxuan Mu, Muhammad~Gohar Javed, Sen Wang, and Li Cheng.
\newblock Momask: Generative masked modeling of 3d human motions.
\newblock In \emph{Proceedings of the IEEE/CVF Conference on Computer Vision and Pattern Recognition}, pages 1900--1910, 2024{\natexlab{a}}.

\bibitem[Guo et~al.(2024{\natexlab{b}})Guo, Mu, Zuo, Dai, Yan, Lu, and Cheng]{guo2024generative}
Chuan Guo, Yuxuan Mu, Xinxin Zuo, Peng Dai, Youliang Yan, Juwei Lu, and Li Cheng.
\newblock Generative human motion stylization in latent space.
\newblock \emph{arXiv preprint arXiv:2401.13505}, 2024{\natexlab{b}}.

\bibitem[Guo et~al.(2025)Guo, Qu, Rahmani, Soh, Hu, Ke, and Liu]{guo2025tstmotion}
Ziyan Guo, Haoxuan Qu, Hossein Rahmani, Dewen Soh, Ping Hu, Qiuhong Ke, and Jun Liu.
\newblock Tstmotion: Training-free scene-aware text-to-motion generation.
\newblock \emph{arXiv preprint arXiv:2505.01182}, 2025.

\bibitem[Harvey et~al.(2020)Harvey, Yurick, Nowrouzezahrai, and Pal]{harvey2020robust}
F{\'e}lix~G Harvey, Mike Yurick, Derek Nowrouzezahrai, and Christopher Pal.
\newblock Robust motion in-betweening.
\newblock \emph{ACM Transactions on Graphics (TOG)}, 39\penalty0 (4):\penalty0 60--1, 2020.

\bibitem[Ho and Salimans(2022)]{ho2022classifier}
Jonathan Ho and Tim Salimans.
\newblock Classifier-free diffusion guidance.
\newblock \emph{arXiv preprint arXiv:2207.12598}, 2022.

\bibitem[Ho et~al.(2020)Ho, Jain, and Abbeel]{ho2020denoising}
Jonathan Ho, Ajay Jain, and Pieter Abbeel.
\newblock Denoising diffusion probabilistic models.
\newblock \emph{Advances in neural information processing systems}, 33:\penalty0 6840--6851, 2020.

\bibitem[Jang et~al.(2022)Jang, Park, and Lee]{jang2022motion}
Deok-Kyeong Jang, Soomin Park, and Sung-Hee Lee.
\newblock Motion puzzle: Arbitrary motion style transfer by body part.
\newblock \emph{ACM Transactions on Graphics (TOG)}, 41\penalty0 (3):\penalty0 1--16, 2022.

\bibitem[Jiang et~al.(2023)Jiang, Chen, Liu, Yu, Yu, and Chen]{jiang2023motiongpt}
Biao Jiang, Xin Chen, Wen Liu, Jingyi Yu, Gang Yu, and Tao Chen.
\newblock Motiongpt: Human motion as a foreign language.
\newblock \emph{Advances in Neural Information Processing Systems}, 36:\penalty0 20067--20079, 2023.

\bibitem[Jiang et~al.(2025)Jiang, Chen, Zhang, Yin, Li, Yu, and Fan]{jiang2025motionchain}
Biao Jiang, Xin Chen, Chi Zhang, Fukun Yin, Zhuoyuan Li, Gang Yu, and Jiayuan Fan.
\newblock Motionchain: Conversational motion controllers via multimodal prompts.
\newblock In \emph{European Conference on Computer Vision}, pages 54--74. Springer, 2025.

\bibitem[Karunratanakul et~al.(2023)Karunratanakul, Preechakul, Suwajanakorn, and Tang]{karunratanakul2023guided}
Korrawe Karunratanakul, Konpat Preechakul, Supasorn Suwajanakorn, and Siyu Tang.
\newblock Guided motion diffusion for controllable human motion synthesis.
\newblock In \emph{Proceedings of the IEEE/CVF International Conference on Computer Vision}, pages 2151--2162, 2023.

\bibitem[Kim et~al.(2023)Kim, Kim, and Choi]{kim2023flame}
Jihoon Kim, Jiseob Kim, and Sungjoon Choi.
\newblock Flame: Free-form language-based motion synthesis \& editing.
\newblock In \emph{Proceedings of the AAAI Conference on Artificial Intelligence}, pages 8255--8263, 2023.

\bibitem[Li et~al.(2024)Li, Chibane, He, Pearl, Geiger, and Pons-Moll]{li2024unimotion}
Chuqiao Li, Julian Chibane, Yannan He, Naama Pearl, Andreas Geiger, and Gerard Pons-Moll.
\newblock Unimotion: Unifying 3d human motion synthesis and understanding.
\newblock \emph{arXiv preprint arXiv:2409.15904}, 2024.

\bibitem[Lin et~al.(2023)Lin, Zeng, Lu, Cai, Zhang, Wang, and Zhang]{lin2023motion}
Jing Lin, Ailing Zeng, Shunlin Lu, Yuanhao Cai, Ruimao Zhang, Haoqian Wang, and Lei Zhang.
\newblock Motion-x: A large-scale 3d expressive whole-body human motion dataset.
\newblock \emph{Advances in Neural Information Processing Systems}, 36:\penalty0 25268--25280, 2023.

\bibitem[Ling et~al.(2024)Ling, Han, Li, Shen, Cheng, and Zou]{ling2024motionllama}
Zeyu Ling, Bo Han, Shiyang Li, Hongdeng Shen, Jikang Cheng, and Changqing Zou.
\newblock Motionllama: A unified framework for motion synthesis and comprehension.
\newblock \emph{arXiv preprint arXiv:2411.17335}, 2024.

\bibitem[Lipman et~al.(2022)Lipman, Chen, Ben-Hamu, Nickel, and Le]{lipman2022flow}
Yaron Lipman, Ricky~TQ Chen, Heli Ben-Hamu, Maximilian Nickel, and Matt Le.
\newblock Flow matching for generative modeling.
\newblock \emph{arXiv preprint arXiv:2210.02747}, 2022.

\bibitem[Liu et~al.(2022)Liu, Gong, and Liu]{liu2022flow}
Xingchao Liu, Chengyue Gong, and Qiang Liu.
\newblock Flow straight and fast: Learning to generate and transfer data with rectified flow.
\newblock \emph{arXiv preprint arXiv:2209.03003}, 2022.

\bibitem[Loper et~al.(2023)Loper, Mahmood, Romero, Pons-Moll, and Black]{loper2023smpl}
Matthew Loper, Naureen Mahmood, Javier Romero, Gerard Pons-Moll, and Michael~J Black.
\newblock Smpl: A skinned multi-person linear model.
\newblock In \emph{Seminal Graphics Papers: Pushing the Boundaries, Volume 2}, pages 851--866. 2023.

\bibitem[Lu et~al.(2023)Lu, Chen, Zeng, Lin, Zhang, Zhang, and Shum]{lu2023humantomato}
Shunlin Lu, Ling-Hao Chen, Ailing Zeng, Jing Lin, Ruimao Zhang, Lei Zhang, and Heung-Yeung Shum.
\newblock Humantomato: Text-aligned whole-body motion generation.
\newblock \emph{arXiv preprint arXiv:2310.12978}, 2023.

\bibitem[Luo et~al.(2024)Luo, Hou, Chang, Liu, Wang, and Shan]{luo2024m}
Mingshuang Luo, Ruibing Hou, Hong Chang, Zimo Liu, Yaowei Wang, and Shiguang Shan.
\newblock M3gpt: An advanced multimodal, multitask framework for motion comprehension and generation.
\newblock \emph{arXiv preprint arXiv:2405.16273}, 2024.

\bibitem[Ma et~al.(2024)Ma, Goldstein, Albergo, Boffi, Vanden-Eijnden, and Xie]{ma2024sit}
Nanye Ma, Mark Goldstein, Michael~S Albergo, Nicholas~M Boffi, Eric Vanden-Eijnden, and Saining Xie.
\newblock Sit: Exploring flow and diffusion-based generative models with scalable interpolant transformers.
\newblock \emph{arXiv preprint arXiv:2401.08740}, 2024.

\bibitem[Peebles and Xie(2023)]{peebles2023scalable}
William Peebles and Saining Xie.
\newblock Scalable diffusion models with transformers.
\newblock In \emph{Proceedings of the IEEE/CVF International Conference on Computer Vision}, pages 4195--4205, 2023.

\bibitem[Petrovich et~al.(2021)Petrovich, Black, and Varol]{petrovich2021action}
Mathis Petrovich, Michael~J Black, and G{\"u}l Varol.
\newblock Action-conditioned 3d human motion synthesis with transformer vae.
\newblock In \emph{Proceedings of the IEEE/CVF International Conference on Computer Vision}, pages 10985--10995, 2021.

\bibitem[Pinyoanuntapong et~al.(2024)Pinyoanuntapong, Wang, Lee, and Chen]{pinyoanuntapong2024mmm}
Ekkasit Pinyoanuntapong, Pu Wang, Minwoo Lee, and Chen Chen.
\newblock Mmm: Generative masked motion model.
\newblock In \emph{Proceedings of the IEEE/CVF Conference on Computer Vision and Pattern Recognition}, pages 1546--1555, 2024.

\bibitem[Plappert et~al.(2016)Plappert, Mandery, and Asfour]{plappert2016kit}
Matthias Plappert, Christian Mandery, and Tamim Asfour.
\newblock The kit motion-language dataset.
\newblock \emph{Big data}, 4\penalty0 (4):\penalty0 236--252, 2016.

\bibitem[Polyak et~al.(2024)Polyak, Zohar, Brown, Tjandra, Sinha, Lee, Vyas, Shi, Ma, Chuang, Yan, Choudhary, Wang, Sethi, Pang, Ma, Misra, Hou, Wang, Jagadeesh, Li, Zhang, Singh, Williamson, Le, Yu, Singh, Zhang, Vajda, Duval, Girdhar, Sumbaly, Rambhatla, Tsai, Azadi, Datta, Chen, Bell, Ramaswamy, Sheynin, Bhattacharya, Motwani, Xu, Li, Hou, Hsu, Yin, Dai, Taigman, Luo, Liu, Wu, Zhao, Kirstain, He, He, Pumarola, Thabet, Sanakoyeu, Mallya, Guo, Araya, Kerr, Wood, Liu, Peng, Vengertsev, Schonfeld, Blanchard, Juefei-Xu, Nord, Liang, Hoffman, Kohler, Fire, Sivakumar, Chen, Yu, Gao, Georgopoulos, Moritz, Sampson, Li, Parmeggiani, Fine, Fowler, Petrovic, and Du]{polyak2024moviegencastmedia}
Adam Polyak, Amit Zohar, Andrew Brown, Andros Tjandra, Animesh Sinha, Ann Lee, Apoorv Vyas, Bowen Shi, Chih-Yao Ma, Ching-Yao Chuang, David Yan, Dhruv Choudhary, Dingkang Wang, Geet Sethi, Guan Pang, Haoyu Ma, Ishan Misra, Ji Hou, Jialiang Wang, Kiran Jagadeesh, Kunpeng Li, Luxin Zhang, Mannat Singh, Mary Williamson, Matt Le, Matthew Yu, Mitesh~Kumar Singh, Peizhao Zhang, Peter Vajda, Quentin Duval, Rohit Girdhar, Roshan Sumbaly, Sai~Saketh Rambhatla, Sam Tsai, Samaneh Azadi, Samyak Datta, Sanyuan Chen, Sean Bell, Sharadh Ramaswamy, Shelly Sheynin, Siddharth Bhattacharya, Simran Motwani, Tao Xu, Tianhe Li, Tingbo Hou, Wei-Ning Hsu, Xi Yin, Xiaoliang Dai, Yaniv Taigman, Yaqiao Luo, Yen-Cheng Liu, Yi-Chiao Wu, Yue Zhao, Yuval Kirstain, Zecheng He, Zijian He, Albert Pumarola, Ali Thabet, Artsiom Sanakoyeu, Arun Mallya, Baishan Guo, Boris Araya, Breena Kerr, Carleigh Wood, Ce Liu, Cen Peng, Dimitry Vengertsev, Edgar Schonfeld, Elliot Blanchard, Felix Juefei-Xu, Fraylie Nord, Jeff Liang, John Hoffman, Jonas
  Kohler, Kaolin Fire, Karthik Sivakumar, Lawrence Chen, Licheng Yu, Luya Gao, Markos Georgopoulos, Rashel Moritz, Sara~K. Sampson, Shikai Li, Simone Parmeggiani, Steve Fine, Tara Fowler, Vladan Petrovic, and Yuming Du.
\newblock Movie gen: A cast of media foundation models, 2024.

\bibitem[Qin et~al.(2022)Qin, Zheng, and Zhou]{qin2022motion}
Jia Qin, Youyi Zheng, and Kun Zhou.
\newblock Motion in-betweening via two-stage transformers.
\newblock \emph{ACM Trans. Graph.}, 41\penalty0 (6):\penalty0 184--1, 2022.

\bibitem[Radford et~al.(2021)Radford, Kim, Hallacy, Ramesh, Goh, Agarwal, Sastry, Askell, Mishkin, Clark, et~al.]{radford2021learning}
Alec Radford, Jong~Wook Kim, Chris Hallacy, Aditya Ramesh, Gabriel Goh, Sandhini Agarwal, Girish Sastry, Amanda Askell, Pamela Mishkin, Jack Clark, et~al.
\newblock Learning transferable visual models from natural language supervision.
\newblock In \emph{International conference on machine learning}, pages 8748--8763. PMLR, 2021.

\bibitem[Sampieri et~al.(2024)Sampieri, Palma, Spinelli, and Galasso]{sampieri2024length}
Alessio Sampieri, Alessio Palma, Indro Spinelli, and Fabio Galasso.
\newblock Length-aware motion synthesis via latent diffusion.
\newblock \emph{arXiv preprint arXiv:2407.11532}, 2024.

\bibitem[Shafir et~al.(2023)Shafir, Tevet, Kapon, and Bermano]{shafir2023human}
Yonatan Shafir, Guy Tevet, Roy Kapon, and Amit~H Bermano.
\newblock Human motion diffusion as a generative prior.
\newblock \emph{arXiv preprint arXiv:2303.01418}, 2023.

\bibitem[Shrestha et~al.(2025)Shrestha, Liu, Ros, Yuan, and Fern]{shrestha2025generating}
Aayam Shrestha, Pan Liu, German Ros, Kai Yuan, and Alan Fern.
\newblock Generating physically realistic and directable human motions from multi-modal inputs.
\newblock In \emph{European Conference on Computer Vision}, pages 1--17. Springer, 2025.

\bibitem[Song et~al.(2020)Song, Meng, and Ermon]{song2020denoising}
Jiaming Song, Chenlin Meng, and Stefano Ermon.
\newblock Denoising diffusion implicit models.
\newblock \emph{arXiv preprint arXiv:2010.02502}, 2020.

\bibitem[Song et~al.(2023)Song, Zhang, Yin, Mardani, Liu, Kautz, Chen, and Vahdat]{song2023loss}
Jiaming Song, Qinsheng Zhang, Hongxu Yin, Morteza Mardani, Ming-Yu Liu, Jan Kautz, Yongxin Chen, and Arash Vahdat.
\newblock Loss-guided diffusion models for plug-and-play controllable generation.
\newblock In \emph{International Conference on Machine Learning}, pages 32483--32498. PMLR, 2023.

\bibitem[Song et~al.(2024)Song, Jin, Li, Chen, Hao, Hou, Li, and Qin]{Song_2024_CVPR}
Wenfeng Song, Xingliang Jin, Shuai Li, Chenglizhao Chen, Aimin Hao, Xia Hou, Ning Li, and Hong Qin.
\newblock Arbitrary motion style transfer with multi-condition motion latent diffusion model.
\newblock In \emph{Proceedings of the IEEE/CVF Conference on Computer Vision and Pattern Recognition (CVPR)}, pages 821--830, 2024.

\bibitem[Su et~al.(2024)Su, Ahmed, Lu, Pan, Bo, and Liu]{su2024roformer}
Jianlin Su, Murtadha Ahmed, Yu Lu, Shengfeng Pan, Wen Bo, and Yunfeng Liu.
\newblock Roformer: Enhanced transformer with rotary position embedding.
\newblock \emph{Neurocomputing}, 568:\penalty0 127063, 2024.

\bibitem[Sun et~al.(2024)Sun, Zheng, Huang, Ma, Huang, and Hu]{sun2024lgtm}
Haowen Sun, Ruikun Zheng, Haibin Huang, Chongyang Ma, Hui Huang, and Ruizhen Hu.
\newblock Lgtm: Local-to-global text-driven human motion diffusion model.
\newblock In \emph{ACM SIGGRAPH 2024 Conference Papers}, pages 1--9, 2024.

\bibitem[Tevet et~al.(2022)Tevet, Gordon, Hertz, Bermano, and Cohen-Or]{tevet2022motionclip}
Guy Tevet, Brian Gordon, Amir Hertz, Amit~H Bermano, and Daniel Cohen-Or.
\newblock Motionclip: Exposing human motion generation to clip space.
\newblock In \emph{European Conference on Computer Vision}, pages 358--374. Springer, 2022.

\bibitem[Tevet et~al.(2023)Tevet, Raab, Gordon, Shafir, Cohen-or, and Bermano]{tevet2023human}
Guy Tevet, Sigal Raab, Brian Gordon, Yoni Shafir, Daniel Cohen-or, and Amit~Haim Bermano.
\newblock Human motion diffusion model.
\newblock In \emph{The Eleventh International Conference on Learning Representations}, 2023.

\bibitem[Vaswani(2017)]{vaswani2017attention}
A Vaswani.
\newblock Attention is all you need.
\newblock \emph{Advances in Neural Information Processing Systems}, 2017.

\bibitem[Wang et~al.(2024{\natexlab{a}})Wang, Guo, Huang, Huang, Wang, You, Li, and Zhao]{wang2024frieren}
Yongqi Wang, Wenxiang Guo, Rongjie Huang, Jiawei Huang, Zehan Wang, Fuming You, Ruiqi Li, and Zhou Zhao.
\newblock Frieren: Efficient video-to-audio generation with rectified flow matching.
\newblock \emph{arXiv preprint arXiv:2406.00320}, 2024{\natexlab{a}}.

\bibitem[Wang et~al.(2024{\natexlab{b}})Wang, Huang, Zhang, Ouyang, Jiao, Feng, Zhou, Wan, Tang, and Xu]{wang2024motiongpt}
Yuan Wang, Di Huang, Yaqi Zhang, Wanli Ouyang, Jile Jiao, Xuetao Feng, Yan Zhou, Pengfei Wan, Shixiang Tang, and Dan Xu.
\newblock Motiongpt-2: A general-purpose motion-language model for motion generation and understanding.
\newblock \emph{arXiv preprint arXiv:2410.21747}, 2024{\natexlab{b}}.

\bibitem[Wu et~al.(2024)Wu, Zhao, Wang, Tai, and Tang]{wu2024motionllm}
Qi Wu, Yubo Zhao, Yifan Wang, Yu-Wing Tai, and Chi-Keung Tang.
\newblock Motionllm: Multimodal motion-language learning with large language models.
\newblock \emph{arXiv preprint arXiv:2405.17013}, 2024.

\bibitem[Xie et~al.(2023)Xie, Jampani, Zhong, Sun, and Jiang]{xie2023omnicontrol}
Yiming Xie, Varun Jampani, Lei Zhong, Deqing Sun, and Huaizu Jiang.
\newblock Omnicontrol: Control any joint at any time for human motion generation.
\newblock \emph{arXiv preprint arXiv:2310.08580}, 2023.

\bibitem[Yang et~al.(2024)Yang, Su, Zhang, Chen, Qian, Liu, and Gan]{yang2024unimumo}
Han Yang, Kun Su, Yutong Zhang, Jiaben Chen, Kaizhi Qian, Gaowen Liu, and Chuang Gan.
\newblock Unimumo: Unified text, music and motion generation.
\newblock \emph{arXiv preprint arXiv:2410.04534}, 2024.

\bibitem[Zhang et~al.(2023{\natexlab{a}})Zhang, Zhang, Cun, Zhang, Zhao, Lu, Shen, and Shan]{zhang2023generating}
Jianrong Zhang, Yangsong Zhang, Xiaodong Cun, Yong Zhang, Hongwei Zhao, Hongtao Lu, Xi Shen, and Ying Shan.
\newblock Generating human motion from textual descriptions with discrete representations.
\newblock In \emph{Proceedings of the IEEE/CVF conference on computer vision and pattern recognition}, pages 14730--14740, 2023{\natexlab{a}}.

\bibitem[Zhang et~al.(2022)Zhang, Cai, Pan, Hong, Guo, Yang, and Liu]{zhang2022motiondiffuse}
Mingyuan Zhang, Zhongang Cai, Liang Pan, Fangzhou Hong, Xinying Guo, Lei Yang, and Ziwei Liu.
\newblock Motiondiffuse: Text-driven human motion generation with diffusion model.
\newblock \emph{arXiv preprint arXiv:2208.15001}, 2022.

\bibitem[Zhang et~al.(2023{\natexlab{b}})Zhang, Li, Cai, Ren, Yang, and Liu]{zhang2023finemogen}
Mingyuan Zhang, Huirong Li, Zhongang Cai, Jiawei Ren, Lei Yang, and Ziwei Liu.
\newblock Finemogen: Fine-grained spatio-temporal motion generation and editing.
\newblock \emph{Advances in Neural Information Processing Systems}, 36:\penalty0 13981--13992, 2023{\natexlab{b}}.

\bibitem[Zhang et~al.(2025)Zhang, Jin, Gu, Hong, Cai, Huang, Zhang, Guo, Yang, He, et~al.]{zhang2025large}
Mingyuan Zhang, Daisheng Jin, Chenyang Gu, Fangzhou Hong, Zhongang Cai, Jingfang Huang, Chongzhi Zhang, Xinying Guo, Lei Yang, Ying He, et~al.
\newblock Large motion model for unified multi-modal motion generation.
\newblock In \emph{European Conference on Computer Vision}, pages 397--421. Springer, 2025.

\bibitem[Zhao et~al.(2024)Zhao, Shi, Yu, Zhou, and Lu]{zhao2024flowturbo}
Wenliang Zhao, Minglei Shi, Xumin Yu, Jie Zhou, and Jiwen Lu.
\newblock Flowturbo: Towards real-time flow-based image generation with velocity refiner.
\newblock \emph{arXiv preprint arXiv:2409.18128}, 2024.

\bibitem[Zhong et~al.(2025)Zhong, Xie, Jampani, Sun, and Jiang]{zhong2025smoodi}
Lei Zhong, Yiming Xie, Varun Jampani, Deqing Sun, and Huaizu Jiang.
\newblock Smoodi: Stylized motion diffusion model.
\newblock In \emph{European Conference on Computer Vision}, pages 405--421. Springer, 2025.

\bibitem[Zhou and Wang(2023)]{zhou2023ude}
Zixiang Zhou and Baoyuan Wang.
\newblock Ude: A unified driving engine for human motion generation.
\newblock In \emph{Proceedings of the IEEE/CVF Conference on Computer Vision and Pattern Recognition}, pages 5632--5641, 2023.

\bibitem[Zhou et~al.(2023)Zhou, Wan, and Wang]{zhou2023unified}
Zixiang Zhou, Yu Wan, and Baoyuan Wang.
\newblock A unified framework for multimodal, multi-part human motion synthesis.
\newblock \emph{arXiv preprint arXiv:2311.16471}, 2023.

\bibitem[Zhou et~al.(2024)Zhou, Wan, and Wang]{zhou2024avatargpt}
Zixiang Zhou, Yu Wan, and Baoyuan Wang.
\newblock Avatargpt: All-in-one framework for motion understanding planning generation and beyond.
\newblock In \emph{Proceedings of the IEEE/CVF Conference on Computer Vision and Pattern Recognition}, pages 1357--1366, 2024.

\end{thebibliography}
}

\clearpage
\setcounter{page}{1}
\maketitlesupplementary

\section{Details of Rectified Flows}

\begin{figure}[!h]
  \centering
  \includegraphics[width=\linewidth,page=1]{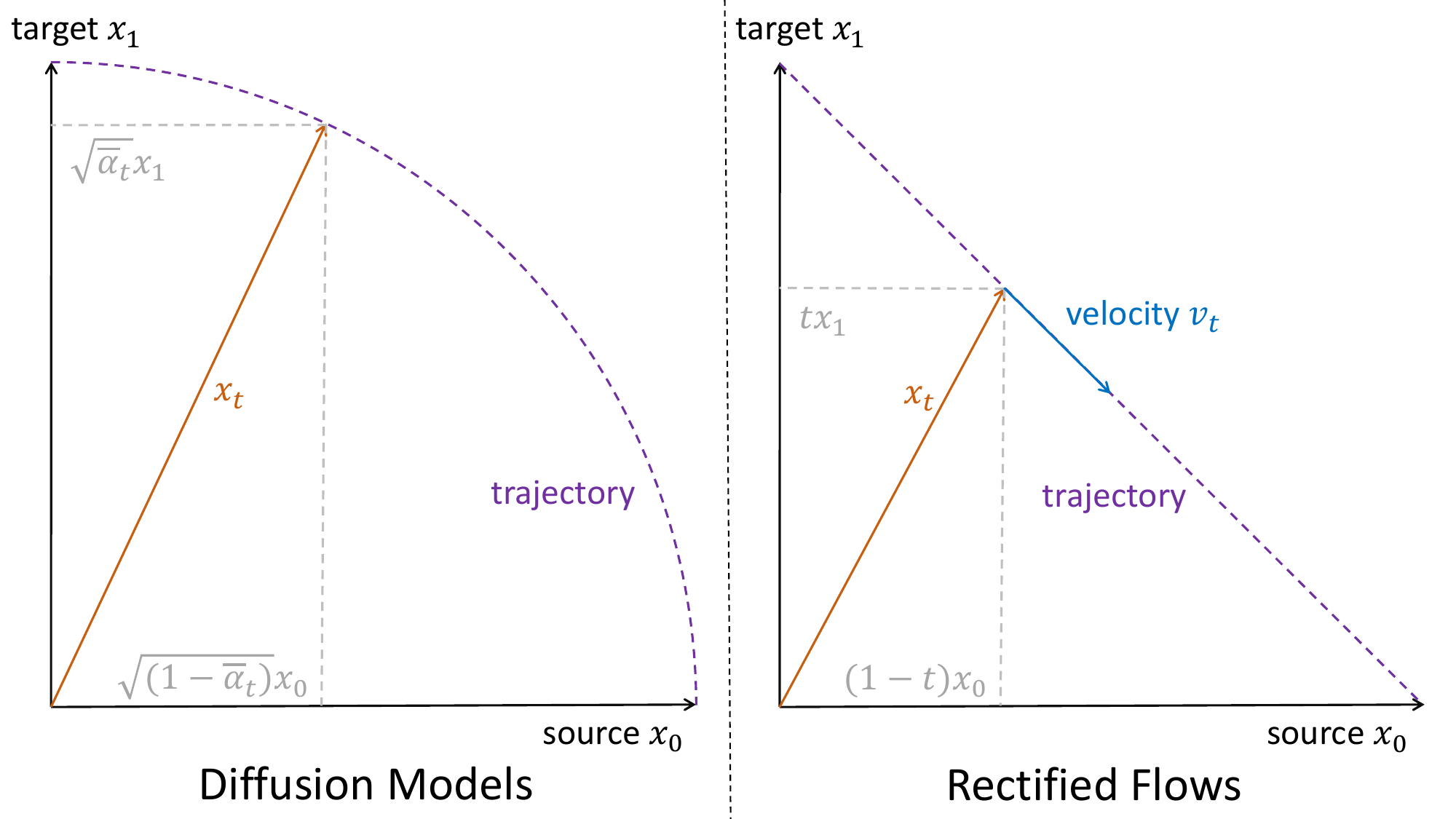}
  \caption{Demonstration of the difference between diffusion models and rectified flows. This difference lies in that the trajectory of diffusion models is based on $x_t = \sqrt{(1-\overline{\alpha_t})} x_0 + \sqrt{\overline{\alpha_t}} \epsilon$, while the trajectory of rectified flows is based on $x_t = (1-t)x_0 + tx_1$. This distinction leads to more robust learning by maintaining a constant velocity, contributing to the model's efficiency \cite{zhao2024flowturbo}. }
  \label{fig:comparison}
\end{figure} 

Since the trajectory $x_t$ from $p_1$ to $p_0$ should be as straight as possible, it can be reformulated as the linear interpolation between $x_0$ and $x_1$, and the velocity field $v_t$ can be treated as a constant, namely:
\begin{align}
& x_t = (1-t)x_0 + tx_1 \\
& v_t = \frac{dx_t}{dt} = \frac{\partial\varphi_t(x_0,x_1,t)}{\partial t} = x_1 - x_0
\end{align}
Therefore, the training objective can be reformulated as:
\begin{align}
\mathcal L_{RF}(\theta) = 
\int^1_0 \mathbb{E}_{(x_0,x_1)\sim(p_0,p_1)}[||v_\theta(t, x_t)-(x_1-x_0)||^2_2]dt
\end{align}
After the training of rectified flows is completed, the transfer from $x_1$ to $x_0$ can be described via the numerical integration of ODE:
\begin{align}
& x_{t-\frac{1}{N}} = x_t - \frac{1}{N}v_\theta(t,x_t)
\end{align}
where $N$ is the discretization number of the interval [0,1].

\section{MotionLab Inference} \label{sec:ucfg}
During inference, Classifier-Free Guidance (CFG) \cite{ho2022classifier} is incorporated for both motion generation and motion editing to boost sampling quality and align conditions and target motion. 

For all motion generation tasks, we generate target motion $M_T$ with the guidance of arbitrary conditions $C$:
\begin{align}
\notag
v_\theta(M_T,t,C) = 
& v_\theta(M_T|t,\emptyset)+ \\
&\lambda_C[v_\theta(M_T|t,C)-v_\theta(M_T|t,\emptyset)]
\end{align}
where $t$ is the timestep and $\lambda_C>1$ is a hyper-parameter to control the strength of the corresponding conditional guidance.

For all motion editing tasks, which aim to modify the source motion based on the condition. Hence, we generate the target motion $M_T$ with source motion $M_S$ first and then condition $C$:
\begin{align}
\notag
v_\theta(M_T,t,M_S,C) = & v_\theta(M_T|t,\emptyset,\emptyset) \\
& +\lambda_S[v_\theta(M_T|t,S,\emptyset)-v_\theta(M_T|t,\emptyset,\emptyset)] \nonumber \\ 
&+\lambda_C[v_\theta(M_T|t,S,C)-v_\theta(M_T|t,S,\emptyset)]
\end{align}
where $\lambda_S>1$ is a hyper-parameter to control the strength of source motion guidance.

\section{Memory Usage and Time Cost}
The maximum memory usage during training is 23 GB for each GPU. The memory usage and the time spent during inference are summarized in the following Table~\ref{tab:ablation_MUTC}.
\begin{table}[h]
\resizebox{\linewidth}{!}{
\begin{tabular}{ccccccc}
\toprule
Metric & text gen & traj. gen & text edit & traj. edit & in-between & style transfer \\ \hline
memory usage (GB) & 4.16 & 4.31 & 5.83 & 6.81 & 4.32 & 5.74 \\
time spend (AITS) & 0.068 & 0.134 & 0.160 & 0.191 & 0.142 & 0.152 \\
\bottomrule
\end{tabular}
}
\caption{The memory usage and time cost of MotionLab.}
\label{tab:ablation_MUTC}
\end{table}

\section{Additional Quantitative Results}
\label{sec:add_quantitative}

As shown in Table~\ref{tab:inbetween}, our framework outperforms CondMDI on all settings, illustrating the effectiveness of our framework in motion in-between. 

\begin{table}[!h]
\resizebox{\linewidth}{!}{
\begin{tabular}{ccccccc}
\hline
Method & Frames & FID$\downarrow$ & \makecell{R-precision\\Top-3$\uparrow$}  & Diversity$\rightarrow$ &  \makecell{Foot skating\\ratio$\downarrow$} & \makecell{Keyframe\\error$\downarrow$} \\ \hline
\multirow{3}{*}{CondMDI \cite{cohan2024flexible}} & 1 & 0.1551 & 0.6787 & 9.5807 & 0.0936 & 0.3739 \\ 
 & 5 & 0.1731 & 0.6823 & 9.3053 & 0.0850 & 0.1789 \\
 & 20 & 0.2253 & 0.6821 & 9.1151 & 0.0806 & 0.0754 \\ \hline
\multirow{3}{*}{Ours} & 1 & 0.7547 & 0.6681 & 8.9058 & 0.0779 & 0.0875 \\ 
 & 5 & 0.0724 & 0.9146 & 9.4406 & 0.0504 & 0.0283 \\
 & 20 & 0.0288 & 0.9914 & 9.5447 & 0.0216 & 0.0215 \\ \hline
\end{tabular}
}
\caption{Evaluation of motion in-between with CondMDI \cite{cohan2024flexible} on HumanML3D \cite{Guo_2022_CVPR} dataset.}
\label{tab:inbetween}
\vspace{-5mm}
\end{table}

\begin{figure}[!h]
  \centering
  \includegraphics[width=\linewidth]{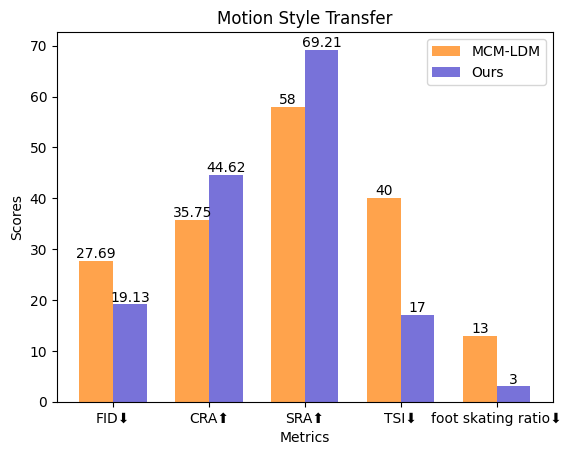}
  \caption{Comparison of the motion style transfer with MCM-LDM \cite{Song_2024_CVPR} on a subset of HumanML3D \cite{Guo_2022_CVPR}. This shows that our model has a stronger ability to preserve the semantics of source motion and a stronger ability to learn the style of style motion.}
  \label{fig:style}
\end{figure}

Also, as shown in Figure~\ref{fig:style}, our framework also outperforms MCM-LDM on all metrics, demonstrating the effectiveness of our framework in motion style transfer.

\section{Additional Ablation Studies}
\label{sec:add_ablation}
To further validate the designs in our framework, we perform traditional ablation studies in this section.

\begin{figure*}[]
  \centering
  \includegraphics[width=\linewidth]{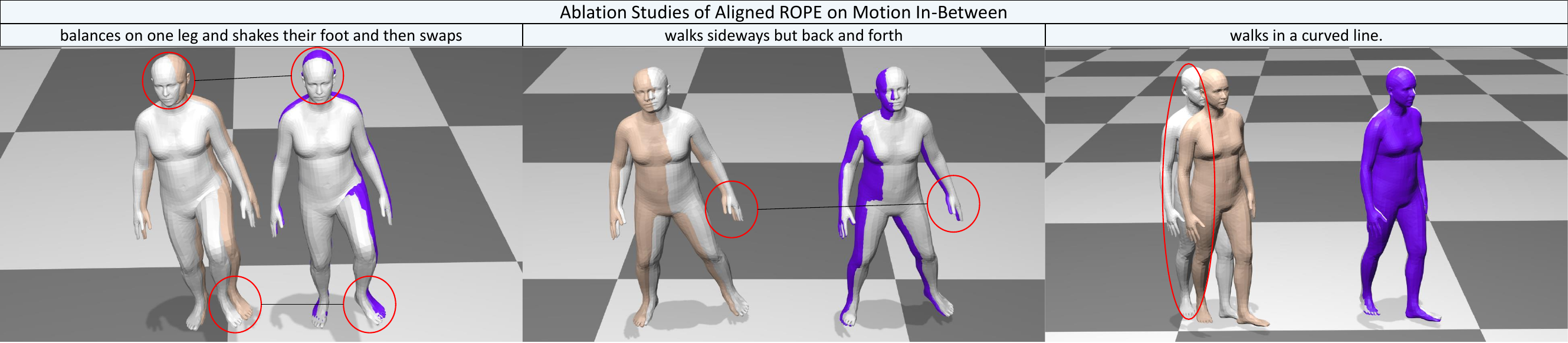}
  \caption{Ablation results of MotionLab on the motion in-between (with text). Beige motion is use 1D-learnable position encoding, purple motion use Aligned ROPE, and gray motions are the poses provided in keyframes, demonstrating the importance of Aligned ROPE.}
  \label{fig:quantitive_inbetween}
\end{figure*}

To further validate the Aligned ROPE, we also introduce the variant of 3D-Learnable and 3D-ROPE to distinguish the source motion, target motion, and trajectory. As shown in Table~\ref{tab:ablation_rope} and Figure~\ref{fig:quantitive_inbetween}, 1D-position encoding is better than 3D-position encoding by avoiding introducing distances between different modalities, and ROPE are better than learnable position encoding by explicit positional encoding. Hence, our 1D-ROPE outperforms all other variants, demonstrating its effectiveness in embedding the position information into tokens.

To further validate the motion curriculum learning, we adopt three variants: removing the masked pre-training and directly supervised fine-tuning in order; with masked pre-training but supervised fine-tuning all tasks together; and introducing masked reconstruction, motion in-between, and trajectory-based motion generation in an orderly manner. As shown in Table~\ref{tab:ablation_learning}, motion curriculum learning outperforms all other variants, highlighting the effectiveness of masked pre-training and fine-tuning tasks in order to avoid gradient conflicts between different tasks. Specifically, the variant of masked pre-training in order to demonstrate the necessity of introducing motion in-between and trajectory-based motion generation together, or it will greatly weaken the performance of the model in the latter task.

The explanation of ``w/o task instruction modulation'' uses 'null' as the instruction for all tasks, rather than learned task tokens or one-hot encoding vectors. We have conducted an additional ablation experiment to examine these situations, which can be suboptimal due to the random initialization of their parameters, as shown in Table~\ref{tab:ablation_TIM}.

\begin{table}[!h]
\resizebox{\linewidth}{!}{
\begin{tabular}{cccccccc}
\toprule
Method & text gen. (FID) & traj. gen. (avg. err.) & text edit (R@1) & traj. edit (R@1) & in-between (avg. err.) & style transfer (CRA) & style transfer (SRA) \\ \hline
w/o task instruction modulation & 0.223 & 0.0401 & 55.96 & 70.01 & 0.0288 & 40.55 & 63.91 \\
one-hot encoding & 0.187 & 0.0369 & 56.18 & 71.52 & 0.0287 & 43.20 & 66.98 \\
learnable tokens & 0.183 & 0.0356 & 56.03 & 71.89 & 0.0288 & 41.20 & 64.98 \\
\hline
\textbf{Ours} & \textbf{0.167} & \textbf{0.0334} & \textbf{56.34} & \textbf{72.65} & \textbf{0.0283} & \textbf{44.62} & \textbf{69.21} \\
\bottomrule
\end{tabular}
}
\caption{Ablation studies of Task Instruction Modulation.}
\label{tab:ablation_TIM}
\end{table}

To further validate the choice and combinations of the tasks, we also introduce the variants of different tasks. As shown in Table~\ref{tab:ablation_task}, improper combination of tasks will cause the unified framework to be weaker than the ours specialist models, while our carefully selected combination of all tasks makes our unified framework beat ours specialist models.

\begin{table*}[!h]
\resizebox{\linewidth}{!}{
\begin{tabular}{cccccccc}
\hline
Method & text gen. (FID) & traj. gen. (avg. err.) & text edit (R@1) & traj. edit (R@1) & in-between (avg. err.) & style transfer (CRA) & style transfer (SRA) \\ \hline
1D-Learnable & 0.246 & 0.0886 & 45.39 & 61.99 & 0.0756 & 39.40 & 56.59 \\
3D-Learnable & 0.346 & 0.1865 & 35.46 & 53.74 & 0.1460 & 36.99 & 58.81  \\
3D-ROPE & 0.241 & 0.0579 & 51.34 & 70.00 & 0.0354 & 42.96 & 62.46 \\
1D-ROPE (ours) & 0.167 & 0.0334 & 56.34 & 72.65 & 0.0273 & 44.62 & 69.21 \\
\hline
\end{tabular}
}
\caption{Ablation studies of our MotionLab's position encoding on each task.}
\label{tab:ablation_rope}
\end{table*}

\begin{table*}[!h]
\resizebox{\linewidth}{!}{
\begin{tabular}{cccccccc}
\hline
Method & text gen. (FID) & traj. gen. (avg. err.) & text edit (R@1) & traj. edit (R@1) & in-between (avg. err.) & style transfer (CRA) & style transfer (SRA) \\ \hline
random selection based on FID & 2.236 & 0.1983 & 28.56 & 36.61 & 0.1682 & 26.61 & 34.23 \\
removing the masked pre-training & 0.861 & 0.0932 & 44.99 & 63.92 & 0.0639 & 39.63 & 57.59  \\
supervised fine-tuning all tasks together & 1.331 & 0.1317 & 38.19 & 55.22 & 0.1143 & 36.60 & 50.59 \\
masked pre-training in order & 0.256 & 0.0423 & 56.33 & 69.31 & 0.0264 & 42.67 & 64.39 \\
motion curriculum learning (ours) & 0.167 & 0.0334 & 56.34 & 72.65 & 0.0273 & 44.62 & 69.21 \\
\hline
\end{tabular}
}
\caption{Ablation studies of our MotionLab's motion curriculum learning on each task.}
\label{tab:ablation_learning}
\end{table*}

\begin{table*}[!h]
\resizebox{\linewidth}{!}{
\begin{tabular}{cccccc|ccccccc}
\hline
\multicolumn{6}{c|}{Task} & \multicolumn{7}{c}{Metric} \\ \hline 
text gen. & traj. gen & text edit & traj. edit & in-between & style transfer  & text gen. (FID) & traj. gen. (avg. err.) & text edit (R@1) & traj. edit (R@1) & in-between (avg. err.) & style transfer (CRA) & style transfer (SRA) \\ \hline
\multicolumn{6}{c|}{ours specialist models} & 0.209 & 0.0398 & 41.44 & 59.86 & 0.0371 & 43.53 & 67.55 \\ \hline
\ding{51} & $\times$ & $\times$ & $\times$ & $\times$ & \ding{51} & 0.240 & - & - & - & - & 41.23 & 65.53 \\
\ding{51} & $\times$ & \ding{51} & $\times$ & $\times$ & $\times$ & 0.235 & - & 52.79 & - & - & - & - \\
\ding{51} & \ding{51} & $\times$ & $\times$ & \ding{51} & $\times$ & 0.176 & 0.0364 & - & - & 0.0297 & - & - \\
\ding{51} & \ding{51} & \ding{51} & \ding{51} & \ding{51} & $\times$ & 0.171 & 0.0344 & 55.10 & 72.20 & 0.0287 & - & - \\
\ding{51} & \ding{51} & \ding{51} & \ding{51} & \ding{51} & \ding{51} & 0.167 & 0.0334 & 56.34 & 72.65 & 0.0273 & 44.62 & 69.21 \\
\hline
\end{tabular}
}
\caption{Ablation studies of our MotionLab's task combinations.}
\label{tab:ablation_task}
\end{table*}

\section{Representation for Each Modality}
We represent the features of all modalities as tokens for the attention mechanism \cite{vaswani2017attention}. Specifically, source motion and target motion are represented as $M_S \in \mathbb{R}^{N\times D}$ and $M_T \in \mathbb{R}^{N\times D}$, and we first ignore timestep $t$ here. For the instruction, it is represented as $I\in\mathbb{R}^{1\times768}$ extracted from the CLIP \cite{radford2021learning}. For available conditions $C$, the text is represent as $p\in\mathbb{R}^{77\times768}$ extracted from the last hidden layer of CLIP, the trajectory is represented as $h\in \mathbb{R}^{N\times J\times3}$, and the style is represented as $s\in\mathbb{R}^{1\times512}$ extracted from \cite{zhong2025smoodi}. 

\section{Instructions for Each Task}

\begin{table}[!h]
\resizebox{\linewidth}{!}{
\begin{tabular}{ll}
\hline
Task & Instruction \\
\hline
unconditional generation & ``reconstruct given masked source motion.'' \\
masked source motion generation & ``reconstruct given masked source motion.'' \\
reconstruct source motion & ``reconstruct given masked source motion.'' \\
trajectory-based generation (without text) & ``generate motion by given trajectory.'' \\
in-between (without text) & ``generate motion by given key frames.'' \\
style-based generation & ``generate motion by given style.'' \\
trajectory-based editing & ``edit source motion by given trajectory.'' \\
text-based editing & ``edit source motion by given text.'' \\
style transfer & ``generate motion by the given style and content.'' \\
in-between (with text) & ``generate motion by given text and key frames.'' \\
trajectory-based generation (with text) & ``generate motion by given text and trajectory.'' \\
text-based generation & ``generate motion by given text.'' \\
\hline
\end{tabular}
}
\caption{Instructions in the Task Instruction Modulations for each task.}
\label{tab:instructions}
\end{table}

As shown in the Table \ref{tab:instructions}, the instructions in the Task Instruction Modulations for each task are presented, which benefits our framework to distinguish different tasks.

\section{Classifier Free Guidance for Each Task}
\begin{table}[!h]
\resizebox{\linewidth}{!}{
\begin{tabular}{lcc}
\hline
Task & Source Motion Guidance & Condition Guidance \\
\hline
trajectory-based generation (without text) & $-$ & 1.5  \\
in-between (without text) & $-$ & 1.5 \\
text-based generation & - & 5.75 \\
style-based generation & - & 1.5 \\
trajectory-based editing (without text) & 2.25 & 2.25 \\
text-based editing & 2.25 & 2.25 \\
style transfer & 1.5 & 1.5 \\
in-between (with text) & $-$ & 1.75 \\
trajectory-based generation (with text) & $-$ & 1.75 \\
trajectory-based editing (with text) & 2 & 2 \\
\hline
\end{tabular}
}
\caption{Strength of classifier free guidance for each task.}
\label{tab:cfg}
\end{table}

As shown in Table~\ref{tab:cfg}, strengths of classifier-free guidance for each task are presented, which contribute to the results' quality during sampling. We conduct ablation experiments based on the hyperparameters provided by the baseline and finally obtain the above hyperparameters.

\section{3D Assets}
We have borrowed some 3D assets for our video and figure from the Internet, including \href{https://sketchfab.com/3d-models/dojo-matrix-drunken-wrestlers-a7902c72cde2447986ff89e13e78a11f}{Dojo Matrix Drunken Wrestlers}, \href{https://sketchfab.com/3d-models/basketball-court-77af6cb6181e4fe7b56bf15035b33422}{Basketball Court}, \href{https://sketchfab.com/3d-models/grandmas-place-02fa0075c38a482187c78ac0eacec214}{Grandma`s Place}, \href{https://sketchfab.com/3d-models/dae-diorama-retake-small-farm-252ad9f2245e47cba4fbb0cfe5eb6445}{DAE Diorama retake – Small farm}, \href{https://sketchfab.com/3d-models/dae-diorama-retake-small-farm-252ad9f2245e47cba4fbb0cfe5eb6445}{DAE Diorama retake – Small farm}, \href{https://www.fab.com/listings/fe784b4e-ab8b-44b7-885d-140b0f81448b}{Japanese Small Shrine Temple 0002}.

\end{document}